%% file: main.tex
\documentclass[twoside]{article}
\usepackage[accepted]{aistats2017}

\usepackage{lmodern}
\usepackage[utf8]{inputenc} 
\usepackage[T1]{fontenc}    
\usepackage{booktabs}       
\usepackage{amsfonts}       
\usepackage{microtype}      
\usepackage{hyperref}
\usepackage{amsmath,amssymb,amsthm,amsfonts,latexsym}        
\usepackage{url}
\usepackage{graphicx}
\usepackage{bm}
\usepackage{subfigure}
\usepackage{color}
\usepackage{amsthm}

\makeatletter
\def\thm@space@setup{%
  \thm@preskip=\parskip \thm@postskip=0pt
}
\makeatother

\newtheorem{theorem}{Theorem}
\newtheorem{lemma}{Lemma}
\newtheorem{assumption}{Assumption}
\newtheorem{definition}{Inequality}

\newenvironment{proof-sketch}{
\proof}{\endproof}

\newcommand{\eat}[1]{} 

\newcommand{\Ct}{\mbox {$\mathcal{C}_{t}$}}
\newcommand{\cE}{\mbox {$\mathcal{E}$}}
\newcommand{\cI}{\mbox {$\mathcal{I}$}}

\newcommand{\vecw}{\mbox {$\mathbf{w}$}}
\newcommand{\vecx}{\mbox {$\mathbf{x}$}}
\newcommand{\vecr}{\mbox {$\mathbf{r}$}}
\newcommand{\vecb}{\mbox {$\mathbf{b}$}}
\newcommand{\vecmu}{\mbox {$\mathbf{\hat{w}}$}}
\newcommand{\vecphi} {\mbox{$\boldsymbol\phi$}}
\newcommand{\hypo}{\mbox {$\mathcal{H}$}}

\newcommand{\Ft}{\mbox {$\mathcal{F}_{t-1}$}}
\newcommand{\vecs}{\mbox {$\mathbf{S}$}}

\newcommand{\kt}{\mbox {$j$}}
\newcommand{\ks}{\mbox {$j_{t}$}}
\newcommand{\ko}{\mbox {$j^{*}_{t}$}}
\newcommand{\CB}{\mbox {$s_{t}$}}
\newcommand{\CBsq}{\mbox {$s_{t}^{2}$}}

\DeclareMathOperator{\p}{Pr}
\DeclareMathOperator{\Tr}{Tr}
\DeclareMathOperator{\nnz}{nnz}
\DeclareMathOperator*{\argmin}{argmin}
\DeclareMathOperator*{\argmax}{argmax}

\let\oldclearpage\clearpage

\begin{document}

%

%

\twocolumn[

\aistatstitle{Horde of Bandits using Gaussian Markov Random Fields}

\aistatsauthor{ Sharan Vaswani \And Mark Schmidt \And Laks V.S. Lakshmanan }
\aistatsaddress{ University of British Columbia} ]

\input{Abstract}
\input{Introduction}
\input{RelatedWork}
\input{Theory}
\input{Algorithms}
\input{Experiments}

\input{Conclusion}
\bibliographystyle{plain}
\bibliography{ref}
\newpage
{
\let\clearpage\relax
\onecolumn
\appendix
\input{Supp-Learning}
\input{Supp-Proofs-Single}
}
\let\clearpage\oldclearpage

\end{document}

%% file: Abstract.tex
\begin{abstract}
The gang of bandits (GOB) model~\cite{cesa2013gang} is a recent contextual bandits framework that shares information between a set of bandit problems, related by a known (possibly noisy) graph. This model is useful in problems like recommender systems where the large number of users makes it vital to transfer information between users. Despite its effectiveness, the existing GOB model can only be applied to small problems due to its quadratic time-dependence on the number of nodes. Existing solutions to combat the scalability issue require an often-unrealistic  clustering assumption. By exploiting a connection to Gaussian Markov random fields (GMRFs), we show that the GOB model can be made to scale to much larger graphs without additional assumptions. In addition, we propose a Thompson sampling algorithm which uses the recent GMRF sampling-by-perturbation technique, allowing it to scale to even larger problems (leading to a ``horde'' of bandits). We give regret bounds and experimental results for GOB with Thompson sampling and epoch-greedy algorithms, indicating that these methods are as good as or significantly better than ignoring the graph or adopting a clustering-based approach. Finally, when an existing graph is not available, we propose a heuristic for learning it on the fly and show promising results. 
\end{abstract}

%% file: Introduction.tex
\vspace*{-4ex}
\section{Introduction}
\label{sec:introduction}
\vspace*{-2ex}
Consider a newly established recommender system (RS) which has little or no information about the users' preferences or any available rating data. The unavailability of rating data implies that we can not use traditional collaborative filtering based methods~\cite{su2009survey}. Furthermore, in the scenario of personalized news recommendation or for recommending trending Facebook posts, the set of available items is not fixed but instead changes continuously. This new RS can recommend  items to the users and observe their ratings to learn their preferences from this feedback (``exploration''). However, in order to retain its users, at the same time it should recommend ``relevant'' items that will be liked by and elicit higher ratings from users (``exploitation''). Assuming each item can be described by its content (like tags describing a news article or video), the contextual bandits framework~\cite{li2010contextual} offers a popular approach for addressing this exploration-exploitation trade-off.

However, this framework assumes that users interact with the RS in an isolated manner, when in fact a RS might have an associated social component. In particular, given the large number of users on such systems, we may be able to learn their preferences more quickly by leveraging the relations between them. One way to use a social network of users to improve recommendations is with the recent gang of bandits (GOB) model~\cite{cesa2013gang}. In particular, the GOB model exploits the homophily effect~\cite{mcpherson2001birds} that suggests  users with similar preferences are more likely to form links in a social network. In other words, user preferences vary smoothly across the social graph and tend to be similar for users connected with each other. This allows us to transfer information between users; we can learn about a user from his or her friends' ratings. However, the existing recommendation algorithm in the GOB framework has a \emph{quadratic} time-dependence on the number of nodes (users) and thus can only be used for a small number of users. Several recent works have tried to improve the scaling of the GOB model by clustering the users into groups~\cite{gentile2014online,nguyen2014dynamic}, but this limits the flexibility of the model and loses the ability to model individual users' preferences.


In this paper, we cast the GOB model in the framework of Gaussian Markov random fields (GMRFs) and show how to exploit this connection to scale it to much larger graphs.
Specifically, we interpret the GOB model as the optimization of a Gaussian likelihood on the users' observed ratings and interpret the user-user graph as the prior inverse-covariance matrix of a GMRF. From this perspective, we can efficiently estimate the users' preferences by performing MAP estimation in a GMRF.
In addition, we propose a Thompson sampling GOB variant that exploits the recent sampling-by-perturbation idea from the GMRF literature~\cite{papandreou2010gaussian} to scale to even larger problems. This idea is fairly general and might be of independent interest in the efficient implementation of other Thompson sampling methods. 
We establish regret bounds (Section~\ref{sec:algorithms}) and provide experimental results (Section~\ref{sec:experiments}) for Thompson sampling as well as an epoch-greedy strategy. These experiments indicate that our methods are as good as or significantly better than approaches which ignore the graph or that cluster the nodes. Finally, 
when the graph of users is not available, we propose a heuristic for learning the graph and user preferences simultaneously in an alternating minimization framework (Appendix A).

%% file: RelatedWork.tex
\vspace*{-3ex}
\section{Related Work}
\label{sec:related-work}
\vspace*{-2ex}
\textbf{Social Regularization}: Using social information to improve recommendations was first introduced by Ma et al.~\cite{ma2011recommender}. They used matrix factorization to fit existing rating data but constrained a user's latent vector to be similar to their friends in the social network. Other methods based on collaborative filtering followed~\cite{rao2015collaborative,delporte2013socially}, 
but these works assume that we already have rating data available. Thus, these methods do not address the exploration-exploitation trade-off faced by a new RS that we consider.

\textbf{Bandits}: The multi-armed bandit problem is a classic approach for trading off exploration and exploitation as we collect data~\cite{lai1985asymptotically}. When features (context) for the ``arms'' are available and changing, it is referred to as the \emph{contextual} bandit problem~\cite{auer2002finite,li2010contextual,chu2011contextual}. The contextual bandit framework is important for the scenario we consider where the set of items available is constantly changing, since the features  allow us to make predictions about items we have never seen before.
Algorithms for the contextual bandits problem include epoch-greedy methods~\cite{langford2008epoch}, those based on upper confidence bounds (UCB)~\cite{chu2011contextual,abbasi2011improved}, and Thompson sampling methods~\cite{agrawal2012thompson}. Note that these standard contextual bandit methods do not model the user-user dependencies that we want to exploit.

Several graph-based methods to model dependencies between the users have been explored in the (non-contextual) multi-armed bandit framework~\cite{caron2012leveraging,mannor2011bandits,alon2014nonstochastic,maillard2014latent}, but the GOB model of Cesa-Bianchi et al.~\cite{cesa2013gang} is the first to exploit the network between users in the contextual bandit framework. 
They proposed a UCB-style algorithm and showed that using the graph leads to lower regret from both a theoretical and practical standpoint.
 However, their algorithm has a time complexity that is quadratic in the number of users. This makes it infeasible for typical RS that have tens of thousands (or even millions) of users.

To scale up the GOB model, several recent works propose to cluster the users and assume that users in the same cluster have the same preferences~\cite{gentile2014online,nguyen2014dynamic}. But this solution loses the ability to model individual users' preferences, and indeed {our experiments indicate that in some applications clustering significantly hurts performance}. In contrast, we want to scale up the original GOB model that learns more fine-grained information in the form of a preference-vector specific to each user. 

Another interesting approach to relax the clustering assumption is to cluster both items and users~\cite{li2016collaborative}, but this only applies if we have a fixed set of items. Some works consider item-item similarities to improve recommendations~\cite{valko2014spectral,kocak2014spectral}, but this again requires a fixed set of items while we are interested in RS where the set of items may constantly be changing.
There has also been work on solving a single bandit problem in a distributed fashion~\cite{korda2016distributed}, but this differs from our approach where we are solving an individual bandit problem on each of the $n$ nodes.  Finally, we note that \emph{all} of the existing graph-based works consider relatively small RS datasets ($\sim 1k$ users), while our proposed algorithms can scale to much larger RS. 

%% file: Theory.tex
\vspace*{-3ex}
\section{Scaling up Gang of Bandits}
\label{sec:methodology}
\vspace{-2ex}
In this section we first describe the general GOB framework, then discuss the relationship to GMRFs, and finally show how this leads to more scalable method. In this paper $\Tr(A)$ denotes the trace of matrix $A$, $A \otimes B$ denotes the Kronecker product of matrices $A$ and $B$, $I_{d}$ is used for the $d$-dimensional identity matrix, and $\text{vec}(A)$ is the stacking of the columns of a matrix $A$ into a vector. 
\vspace{-2ex}
\subsection{Gang of Bandits Framework}
\label{subsec:framework}
\vspace{-2ex}
The contextual bandits framework proceeds in rounds. In each round $t$, a set of items $\Ct$  becomes available. These items could be movies released in a particular week, news articles published on a particular day, or trending stories on Facebook. We assume that $\vert \Ct \vert = K$ for all $t$. We assume that each item $j$ can be described by a context (feature) vector $\vecx_{j} \in \mathbb{R}^d$. We use $n$ as the number of users, and denote the (unknown) ground-truth preference vector for user $i$ as $\vecw_{i}^* \in \mathbb{R}^{d}$. Throughout the paper, we assume there is only a single target user per round. It is straightforward extend our results to multiple target users. 

Given a target user $i_{t}$, our task is to recommend an available item $j_{t} \in \Ct$ to them. User $i_{t}$ then provides feedback on the recommended item $j_{t}$ in the form of a rating $r_{i_{t},j_{t}}$. Based on this feedback, the estimated preference vector for user $i_{t}$ is updated. The recommendation algorithm must trade-off between exploration (learning about the users' preferences) and exploitation (obtaining high ratings). 
We evaluate performance using the notion of \emph{regret}, which is the loss in recommendation performance due to lack of knowledge of user preferences. In particular, the regret $R(T)$ after $T$ rounds is given by: 
\begin{eqnarray}
R(T) = \sum_{t = 1}^{T} \bigg[ \max_{j \in \Ct} (\vecw_{i_{t}}^{*T} \vecx_{j}) - \vecw_{i_{t}}^{*T} \vecx_{j_{t}} \bigg].
\label{eq:regret}
\end{eqnarray}

In our analysis we make the following assumptions:
\begin{assumption} The $\ell_2$-norms of the true preference vectors and item feature vectors are bounded from above. Without loss of generality we'll assume $\vert \vert x_{j} \vert \vert_{2} \leq 1$ for all $j$ and  $\vert \vert \vecw_{i}^{*} \vert \vert_{2} \leq 1$ for all $i$. Also without loss of generality, we assume that the ratings are in the range $[0,1]$. \end{assumption}

\begin{assumption} The true ratings can be given by a linear model~\cite{li2010contextual}, meaning that $r_{i,j} = (\vecw_{i}^*)^{T} \vecx_{j} + \eta_{i,j,t}$ for some noise term $\eta_{i,j,t}$. \end{assumption}
These are standard assumptions in the literature.
We denote the history of observations until round $t$ as $\mathbb{H}_{t-1} = \{ (i_{\tau}, j_{\tau}, r_{ i_{\tau}, j_{\tau} })\}_{\tau = 1,2 \cdots t-1}$ and the union of the set of available items until round $t$  along with their corresponding features as $\mathbb{C}_{t-1}$.  

\begin{assumption} 
The noise $\eta_{i,j,t}$ is conditionally sub-Gaussian~\cite{agrawal2012thompson}\cite{cesa2013gang} with zero mean and bounded variance, meaning that $\mathbb{E}[ \eta_{i,j,t} \;|\; \mathbb{C}_{t-1}, \mathbb{H}_{t-1} ] = 0$ and that there exists a $\sigma > 0$ such that for all $\gamma \in \mathbb{R}$, we have $\mathbb{E}[\exp(\gamma \eta_{i,j,t}) \; |\; \mathbb{H}_{t-1}, \mathbb{C}_{t-1}] \leq \exp(\frac{\gamma^{2} \sigma^{2}}{2} )$. 
\label{ass:ass-noise}
\end{assumption}
This assumption implies that for all $i$ and $j$, the conditional mean is given by $\mathbb{E}[ r_{i,j} | \mathbb{C}_{t-1}, \mathbb{H}_{t-1}] = \vecw_{i}^{*T} \vecx_{j}$ and that the conditional variance satisfies $\mathbb{V}[ r_{i,j} | \mathbb{C}_{t-1}, \mathbb{H}_{t-1}] \leq \sigma^{2}$. 

In the GOB framework, we assume access to a (fixed) graph $G = (\mathcal{V},\mathcal{E})$ of users in the form of a social network (or ``trust graph''). Here, the nodes $\mathcal{V}$ correspond to users, whereas the edges $\mathcal{E}$ correspond to friendships or trust relationships. 
The homophily effect implies that the true user preferences vary smoothly across the graph, so we expect the preferences of users connected in the graph to be close to each other. Specifically, 
\begin{assumption} The true user preferences vary smoothly according to the given graph, in the sense that we have a small value of
\[
\sum_{(i_{1},i_{2}) \in \mathcal{E}} \vert \vert \vecw_{i_{1}}^{*} - \vecw_{i_{2}}^{*} \vert \vert^{2}.
\] 
\end{assumption}
Hence, we assume that the graph acts as a correctly-specified prior on the users' true preferences. Note that this assumption implies that nodes in dense subgraphs will have a higher similarity than those in sparse subgraphs (since they will have a larger number of neighbours).

This assumption is violated in some datasets. For example,  in our experiments we consider one dataset in which the available graph is imperfect, in that user preferences do not seem to vary smoothly across all graph edges. Intuitively, we might think that the GOB model might be harmful in this case (compared to ignoring the graph structure). However, in our experiments, we observe that even in these cases, the GOB approach still lead to results as good as ignoring the graph. 
 
The GOB model~\cite{cesa2013gang} solves a contextual bandit problem for each user, where the mean vectors in the different problems are related according to the Laplacian $L$\footnote{To ensure invertibility, we set $L = L_{G} + I_{n}$ where $L_{G}$ is the normalized graph Laplacian.} of the graph $G$. Let $\vecw_{i,t}$ be the preference vector estimate for user $i$ at round $t$. Let $\vecw_{t}$ and $\vecw^{*}$ $\in \mathbb{R}^{dn}$ (respectively) be the concatenation of the vectors $\vecw_{i,t}$ and $\vecw_{i}^*$ across all users. The GOB model solves the following regression problem to find the mean preference vector estimate at round $t$,
\begin{eqnarray}
{\vecw}_{t} = \argmin_{\vecw} \bigg[ \sum_{i = 1}^{n} \sum_{k \in \mathcal{M}_{i,t}} (\vecw_{i}^{T} \vecx_{k} - r_{i,k})^{2}  \nonumber \\ 
+ \lambda \vecw^{T} (L \otimes I_{d}) \vecw  \bigg],
\label{eq:org-loss}
\end{eqnarray}
where $\mathcal{M}_{i,t}$ is the set of items rated by user $i$ up to round $t$. 
The first term is a data-fitting term and models the observed ratings. The second term is the Laplacian regularization and equal to $\sum_{(i,j) \in \mathcal{E}} \lambda \vert \vert \vecw_{i,t} - \vecw_{j,t} \vert \vert_{2}^{2}$. This term models smoothness across the graph with $\lambda > 0$ giving the strength of this  regularization. Note that the same objective function has also been explored for graph-regularized multi-task learning~\cite{evgeniou2004regularized}. 
\subsection{Connection to GMRFs}
\label{subsec:gmrf}
\vspace*{-2ex}
Unfortunately, the approach of Cesa-Bianchi~\cite{cesa2013gang} for solving~\eqref{eq:org-loss} has a computational complexity of $O(d^2n^2)$. To solve~\eqref{eq:org-loss} more efficiently, we now show that it can be interpreted as performing MAP estimation in a GMRF. This will allow us to apply the GOB model to much larger datasets, and lead to an even more scalable algorithm based on Thompson sampling (Section~\ref{sec:algorithms}). 

Consider the following generative model for the ratings $r_{i,j}$ and the user preference vectors $\vecw_i$,
\[
r_{i,j} \sim \mathcal{N}(\vecw_{i}^{T}\vecx_j, \sigma^2), \quad \vecw \sim  \mathcal{N}(0,(\lambda L \otimes I_{d})^{-1}).
\]
This GMRF model assumes that the ratings $r_{i,j}$ are independent given $\vecw_i$ and $\vecx_j$, which is the standard regression assumption. Under this independence assumption the first term in~\eqref{eq:org-loss} is equal up the negative log-likelihood for all of the observed ratings $\vecr_t$ at time $t$, $\log p(\vecr_{t} \; |\; \vecw, \vecx_t, \sigma)$, up to an additive constant and assuming $\sigma = 1$. Similarly, the negative log-prior $p(\vecw \;|\; \lambda, L)$ in this model gives the second term in~\eqref{eq:org-loss} (again, up to an additive constant that does not depend on $\vecw$). Thus, by Bayes rule minimizing~\eqref{eq:org-loss} is equivalent to maximizing the posterior in this GMRF model.


To characterize the posterior, it is helpful to introduce the notation $\vecphi_{i,j} \in \mathbb{R}^{dn}$ to represent the ``global'' feature vector corresponding to recommending item $j$ to user $i$. In particular, let $\vecphi_{i,j}$ be the concatenation of $n$ $d$-dimensional vectors where the $i^{th}$ vector is equal to $\vecx_j$ and the others are zero. 
The rows of the $t \times dn$ dimensional matrix $\Phi_{t}$ correspond to these ``global'' features for all the recommendations made until time $t$. Under this notation, the posterior $p(\vecw \; | \; \vecr_t, \vecw, \Phi_{t})$ is given by a $ \mathcal{N}(\vecmu_{t},\Sigma_{t}^{-1})$ distribution with $\Sigma_{t} = \frac{1}{\sigma^{2}} \Phi_{t}^{T} \Phi_{t} + \lambda (L \otimes I_{d})$ and $\vecmu_{t} = \frac{1}{\sigma^{2}} \Sigma_{t}^{-1} \vecb_{t}$ with $\vecb_{t} = \Phi_{t}^{T} \vecr_{t}$. 
We can view the approach in~\cite{cesa2013gang} as explicitly constructing the dense $dn \times dn$ matrix $\Sigma_{t}^{-1}$, leading to an ${O}(d^2 n^2)$ memory requirement. A new recommendation at round $t$ is thus equivalent to a rank-$1$ update to $\Sigma_{t}$, and even with the Sherman-Morrison formula this leads to an $O(d^2n^2)$ time requirement for each iteration. 
\vspace{-2 ex}
\subsection{Scalability}
\label{subsec:scalability}
\vspace*{-2ex}
Rather than treating $\Sigma_{t}$ as a general matrix, we propose to exploit its structure to scale up the GOB framework to problems where $n$ is very large. In particular, solving~\eqref{eq:org-loss} corresponds to finding the mean vector of the GMRF, which corresponds to solving the linear system $\Sigma_{t} \vecw = \vecb_{t}$. Since $\Sigma_{t}$ is positive-definite, the linear system can be solved using conjugate gradient~\cite{hestenes1952methods}. Conjugate gradient notably does not require $\Sigma_{t}^{-1}$, but instead uses  matrix-vector products $\Sigma_{t} \mathbf{v} = (\Phi_{t}^T\Phi_{t}) \mathbf{v} + \lambda (L \otimes I_d) \mathbf{v}$ for vectors $\mathbf{v} \in \mathbb{R}^{dn}$. Note that $\Phi_{t}^{T} \Phi_{t}$ is block diagonal and has only $O(nd^2)$ non-zeroes. Hence, $\Phi_t^T\Phi_t\mathbf{v}$ can be computed in $O(nd^2)$ time. For computing $(L \otimes I_{d}) \mathbf{v}$, we use that $(B^{T} \otimes A) \mathbf{v} = \text{vec}(AVB)$, where $V$ is an $n \times d$ matrix such that $\text{vec}(V) = \mathbf{v}$. This implies $(L \otimes I_{d}) \mathbf{v}$ can be written as $VL^{T}$ which can be computed in $O(d \cdot \nnz(L))$  time, where $\nnz(L)$ is the number of non-zeroes in $L$. This approach thus has a memory requirement of $O(nd^2 + \nnz(L))$ and a time complexity of $O(\kappa (nd^2 + d \cdot \nnz(L)))$ per mean estimation. Here, $\kappa$ is the number of conjugate gradient iterations which depends on the condition number of the matrix (we used warm-starting by the solution in the previous round for our experiments, which meant that $\kappa = 5$ was enough for convergence). Thus, the algorithm scales linearly in $n$ and in the number of edges of the network (which tends to be linear in $n$ due to the sparsity of social relationships). This enables us to scale to large networks, of the order of $50$K nodes and millions of edges. 

%% file: Algorithms.tex
\vspace*{-3ex}
\section{Alternative Bandit Algorithms}
\label{sec:algorithms}
\vspace*{-3ex}
The above structure can be used to speed up the mean estimation for any algorithm in the GOB framework. However, the LINUCB-like algorithm in~\cite{cesa2013gang} needs to estimate the confidence intervals $\sqrt{\vecphi_{i,j}^{T} \Sigma_{t}^{-1} \vecphi_{i,j}}$ for each available item $j \in \Ct$. Using the GMRF connection, estimating these requires $O(\vert \Ct \vert \kappa (nd^2 + d \cdot \nnz(L)))$ time since we need solve the linear system with $\vert \Ct \vert$ right-hand sides, one for each available item. But this becomes impractical when the number of available items in each round is large. 

We propose two approaches for mitigating this: first, in this section we adapt the epoch-greedy~\cite{langford2008epoch} algorithm to the GOB framework. Epoch-greedy doesn't require confidence intervals and is thus very scalable, but unfortunately it doesn't achieve the optimal regret of $\tilde{O}(\sqrt{T})$. To achieve the optimal regret, we also propose a GOB variant of Thompson sampling~\cite{li2010contextual}. In this section we further exploit the connection to GMRFs to scale Thompson sampling to even larger problems by using the recent sampling-by-perturbation trick~\cite{papandreou2010gaussian}. This GMRF connection and scalability trick might be of independent interest for Thompson sampling in other large-scale problems. 

\vspace*{-2ex} 
\subsection{Epoch-Greedy}
\label{subsec:EG}
\vspace*{-2ex}
Epoch-greedy~\cite{langford2008epoch} is a variant of the popular $\epsilon$-greedy algorithm that explicitly differentiates between exploration and exploitation rounds. An ``exploration'' round consists of  recommending a random item from $\Ct$ to the target user $i_{t}$. The feedback from these exploration rounds is used to learn $\vecw^{*}$. An ``exploitation'' round consists of choosing the available item $\hat{j_{t}}$ which maximizes the expected rating, $\hat{j_{t}} = \argmax_{j \in \Ct} \hat{\vecw}_{t}^{T} \vecphi_{i_{t},j}$. Epoch-greedy proceeds in epochs, where each epoch $q$ consists of 1 exploration round and $s_{q}$ exploitation rounds. 

\textbf{Scalability:} The time complexity for Epoch-Greedy is dominated by the exploitation rounds that require computing the mean and estimating the expected rating for all the available items. Given the mean vector, this estimation takes $O(d \vert \Ct \vert)$ time. The overall time complexity per exploitation round is thus $O(\kappa (nd^2 + d \cdot \nnz(L)) + d \vert \Ct \vert)$. 

\textbf{Regret:} We assume that we incur a maximum regret of $1$ in an exploration round, whereas the regret incurred in an exploitation round depends on how well we have learned $\vecw^{*}$. The attainable regret is thus proportional to the generalization error for the class of hypothesis functions mapping the context vector to an expected rating~\cite{langford2008epoch}. In our case, the class of hypotheses is a set of linear functions (one for each user) with Laplacian regularization. We characterize the generalization error in the GOB framework in terms of its Rademacher complexity~\cite{maurer2006rademacher}, and use this to bound the expected regret leading to the result below. For ease of exposition in the regret bounds, we suppress the factors that don't depend on either $n$, $L$, $\lambda$ or $T$. The complete bound is stated in the supplementary material (Appendix B).   
{
\begin{theorem}
\label{thm:EG}
Under the additional assumption that $\vert \vert w_t \vert \vert_{2} \leq 1$ for all rounds $t$, the expected regret obtained by epoch-greedy in the GOB framework is given as:
\begin{align*}
R(T) = \tilde{O}\left( n^{1/3} \left( \frac{\Tr(L^{-1})}{\lambda n} \right)^{\frac{1}{3}} T^{\frac{2}{3}} \right)
\end{align*}
\begin{proof-sketch}
Let $\mathcal{H}$ be the class of valid hypotheses of linear functions coupled with Laplacian regularization. Let $Err(q,\mathcal{H})$ be the generalization error for $\hypo$ after obtaining $q$ unbiased samples in the exploration rounds. We adapt Corollary 3.1 from~\cite{langford2008epoch} to our context: 
\begin{lemma}
If $s_{q} = \bigg\lfloor \frac{1}{Err(q,\hypo)} \bigg\rfloor$ and $Q_{T}$ is the smallest $Q$ such that $Q + \sum_{q = 1}^{Q} s_{q} \geq T $, the regret obtained by Epoch-Greedy can be bounded as $R(T) \leq 2 Q_{T}$. 
\label{lemma:EG-l1}
\end{lemma}
We use~\cite{maurer2006rademacher} to bound the generalization error of our class of hypotheses in terms of its empirical Rademacher complexity $\hat{\mathcal{R}}_{q}^{n}(\hypo)$. With probability $1 - \delta$, 
\begin{eqnarray}
Err(q,\hypo) \leq \hat{\mathcal{R}}_{q}^{n}(\hypo) + \sqrt{\frac{9 \ln(2/\delta)}{2q}}.
\label{eq:MTL-gen-main}
\end{eqnarray}
Using Theorem 2 in~\cite{maurer2006rademacher} and Theorem 12 from~\cite{bartlett2003rademacher}, we obtain
\begin{eqnarray}
\hat{\mathcal{R}}_{q}^{n}(\hypo) \leq \frac{2}{\sqrt{q}} \sqrt{\frac{12 Tr(L^{-1})}{\lambda}}.
\label{eq:MTL-RC}
\end{eqnarray}
Using~\eqref{eq:MTL-gen-main} and~\eqref{eq:MTL-RC} we obtain
\begin{eqnarray}
Err(q,\hypo) \leq \frac{\bigg[2 \sqrt{12 Tr(L^{-1})/\lambda}+ \sqrt{\frac{9 \ln(2/\delta)}{2}} \bigg]}{\sqrt{q}}.
\label{eq:gen-err-bound-main}
\end{eqnarray}
The theorem follows from~\eqref{eq:gen-err-bound-main} along with Lemma~\ref{lemma:EG-l1}.  
\end{proof-sketch}
\end{theorem}
The effect of the graph on this regret bound is reflected through the term $\Tr(L^{-1})$. For a connected graph, we have the following upper-bound $\frac{\Tr(L^{-1})}{n} \leq \frac{(1 - 1/n)}{\nu_{2}} + \frac{1}{n}$~\cite{maurer2006rademacher}. Here, $\nu_{2}$ is the second smallest eigenvalue of the Laplacian. The value $\nu_2$ represents the algebraic connectivity of the graph~\cite{fiedler1973algebraic}. For a more connected graph, $\nu_{2}$ is higher, the value of $\frac{\Tr(L^{-1})}{n}$ is lower, resulting in a smaller regret. Note that although this result leads to a sub-optimal dependence on $T$ ($T^{\frac{2}{3}}$ instead of $T^{\frac{1}{2}}$), our experiments incorporate a small modification that gives similar performance to the more-expensive LINUCB.

\subsection{Thompson sampling}
\label{subsec:TS}
\vspace*{-2ex}
A common alternative to LINUCB and Epoch-Greedy is Thompson sampling (TS). At each iteration TS uses a sample $\tilde{\vecw}_{t}$ from the posterior distribution at round $t$, $\tilde{\vecw}_{t} \sim \mathcal{N}(\vecw_{t},\Sigma_{t}^{-1})$. It  then selects the item ${j}_{t}$ based on the obtained sample, ${j}_{t} = \argmax_{j \in \Ct} \tilde{\vecw}_{t}^{T} \vecphi_{i_{t},j}$. We show below that the GMRF connection makes TS scalable, but unlike Epoch-Greedy it also achieves the optimal regret.

\textbf{Scalability:} The conventional approach for sampling from a multivariate Gaussian posterior involves forming the Cholesky factorization of the posterior covariance matrix. But in the GOB model the posterior covariance matrix is a $dn$-dimensional matrix where the fill-in from the Cholesky factorization can lead to a computational complexity of $O(d^{2} n^{2})$. In order to implement Thompson sampling for large values of $n$, we adapt the recent sampling-by-perturbation approach~\cite{papandreou2010gaussian} to our setting, and this allows us to sample from a Gaussian prior and then solve a linear system to sample from the posterior. 

Let $\tilde{\vecw}_{0}$ be a sample from the prior distribution and let $\tilde{\vecr}_{t}$ be the perturbed (with standard normal noise) rating vector at round $t$, meaning that $\tilde{\vecr}_{t} = \vecr_{t} + \mathbf{y}_{t}$ for  $\mathbf{y}_{t} \sim \mathcal{N}(0, I_{t})$. In order to obtain a sample  $\tilde{\vecw}_{t}$ from the posterior, we can solve the  linear system
\begin{eqnarray}
\Sigma_{t} \tilde{\vecw}_{t} = (L \otimes I_{d}) \tilde{\vecw}_{0} + \Phi_{t}^{T} \tilde{\vecr}_{t}.
\label{eq:TS-perturbations}
\end{eqnarray}
Let $S$ be the Cholesky factor of $L$ so that $L = S S^{T}$. 
Note that $L \otimes I_{d} = (S \otimes I_{d}) (S \otimes I_{d})^{T}$. If $\mathbf{z} \sim \mathcal{N}(0,I_{dn})$, we can obtain a sample from the prior by solving $(S \otimes I_{d}) \tilde{\vecw}_{0} = \mathbf{z}$. Since $S$ tends to be sparse (using for example~\cite{davis2005algorithm,kyng2016approximate}), this equation can be solved efficiently using conjugate gradient. We can pre-compute and store $S$ and thus obtain a sample from the prior in time  $O(d \cdot \nnz(L))$. Using that $\Phi_{t}^{T} \tilde{\vecr}_{t} = \vecb_{t} + \Phi_{t}^{T} \mathbf{y}_{t}$ in~\eqref{eq:TS-perturbations} and simplifying we obtain
\begin{eqnarray}
\Sigma_{t} \tilde{\vecw}_{t} = (L \otimes I_{d}) \tilde{\vecw}_{0} + \vecb_{t} + \Phi_{t}^{T} \mathbf{y}_{t}
\label{eq:TS-perturbations-simplified}
\end{eqnarray}
As before, this system can be solved efficiently using conjugate gradient. Note that solving~\eqref{eq:TS-perturbations-simplified} results in an exact sample from the $dn$-dimensional posterior. Computing $\Phi_{t}^{T} \mathbf{y}_{t}$ has a time complexity of $O(dt)$. Thus, this approach is faster than the original GOB framework whenever $t < d n^{2}$. Since we focus on the case of large graphs, this condition will tend to hold in our setting. 

We now describe an alternative method of constructing the right side of~\eqref{eq:TS-perturbations-simplified} that doesn't depend on $t$. Observe that computing $\Phi_{t}^{T} \mathbf{y}_{t}$ 
is equivalent to sampling from the distribution $\mathcal{N}(0,\Phi_{t}^{T} \Phi_{t})$. To sample from this distribution, we maintain the Cholesky factor $P_{t}$ of $\Phi_{t}^{T} \Phi_{t}$. Recall that the matrix $\Phi_{t}^{T} \Phi_{t}$ is block diagonal (one block for every user) for all rounds $t$. Hence, its Cholesky factor $P_{t}$ also has a block diagonal structure and requires $\mathcal{O}(n d^{2})$ storage. In each round, we make a recommendation to a single user and thus make a rank-$1$ update to only one $d \times d$ block of $P_{t}$. This is an order $\mathcal{O}(d^{2})$ operation. Once we have an updated $P_{t}$, sampling from $\mathcal{N}(0,\Phi_{t}^{T} \Phi_{t})$ and constructing the right side of~\eqref{eq:TS-perturbations-simplified} is an $O(n d^2)$ operation. The per-round computational complexity for our TS approach is thus $O( \min\{n d^{2},dt\} + d \cdot nnz(L))$ for forming the right side in~\eqref{eq:TS-perturbations-simplified}, $O(nd^{2} + d \cdot \nnz(L))$ for solving the linear system in~\eqref{eq:TS-perturbations-simplified} as well as for computing the mean,  and $O(d \cdot \vert \Ct \vert)$ for selecting the item. Thus, our proposed approach has a complexity linear in the number of nodes and edges and can scale to large networks. 

\textbf{Regret:} To analyze the regret with TS, observe that TS in the GOB framework is equivalent to solving a single $dn$-dimensional contextual bandit problem, but with a modified prior covariance equal to $(\lambda L \otimes I_{d})^{-1}$ instead of $I_{dn}$. We obtain the result below by following a similar argument to Theorem~1 in~\cite{agrawal2012thompson}. The main challenge in the proof is to make use of the available graph to bound the variance of the arms. We first state the result and then sketch the main differences from the original proof. 

\begin{theorem}
\label{thm:TS}
Under the following additional technical assumptions: (a) $\log(K) < (dn-1) \ln(2)$, (b) $\lambda < dn$, and  (c) $\log \left( \frac{3 + T / \lambda dn}{\delta} \right) \leq \log(KT) \log(T/\delta)$, with probability $1 - \delta$, the regret obtained by Thompson Sampling in the GOB framework is given as:
\begin{align*}
R(T) = \tilde{O}\left(\frac{dn\sqrt{T}}{\sqrt{\lambda}}\sqrt{\log \left( \frac{3 \Tr(L^{-1})}{n} + \frac{\Tr(L^{-1}) T}{\lambda dn^{2}\sigma^{2}} \right)} \right)
\end{align*}
\begin{proof-sketch}
To make the notation cleaner, for the  round $t$ and target user $i_{t}$ under consideration, we use $\kt$ to index the available items. Let the index of the optimal item at round $t$ be $\ko$ whereas the index of the item chosen by our algorithm is denoted $\ks$. Let $s_{t}(\kt)$ be the standard deviation in the estimated rating of item $\kt$ at round $t$. It is given as $s_{t}(\kt) = \sqrt{\vecphi_{\kt}^{T} \Sigma_{t-1}^{-1} \vecphi_{\kt}}$. Further, let $l_{t} = \sqrt{dn \log \left( \frac{3 + t / \lambda dn}{\delta} \right)} + \sqrt{3 \lambda}$. Let $\cE^{\mu}(t)$ be the event such that for all $\kt$, 
\begin{align}
\cE^{\mu}(t): \;\; \vert \langle {\vecw}_{t}, \vecphi_{\kt} \rangle - \langle \vecw^{*}, \vecphi_{\kt} \rangle \vert \leq l_{t} s_{t}(\kt) \nonumber
\end{align}
We prove that, for $\delta \in (0,1)$, $\p(\cE^{\mu}(t)) \geq 1 - \delta$. Define $g_{t} = \sqrt{4 \log(tK)} \rho_{t} + l_{t}$, where $\rho_{t}=\sqrt{9d \log \left(\frac{t}{\delta}\right)}$. Let $\gamma = \frac{1}{4e\sqrt{\pi}}$. Given that the event $\cE^{\mu}(t)$ holds with high probability, we follow an argument similar to Lemma 4 of~\cite{agrawal2012thompson} and obtain the following bound:
\begin{align}
R(T) \leq \frac{3 g_{T}}{\gamma} \sum_{t = 1}^{T} s_{t}(\ks) + \frac{2 g_{T}}{\gamma} \sum_{t = 1}^{T} \frac{1}{t^2}  \nonumber \\
+  \frac{6g_{T}}{\gamma} \sqrt{2T \ln{2/\delta}}
\label{eq:regret-bound}
\end{align}
To bound the variance of the selected items, $\sum_{t = 1}^{T} s_{t}(\ks)$, we extend the analysis in~\cite{dani2008stochastic, wen2015efficient} to include the prior covariance term. We thus obtain the following inequality:
\begin{align}
\sum_{t = 1}^{T} s_{t}(\ks) & \leq \sqrt{dn T} \nonumber \\
& \times \sqrt{C \log \left( \frac{\Tr(L^{-1})}{n} \right) + \log \left(3 + \frac{T}{\lambda dn \sigma^{2}} \right)}
\end{align}
where $C = \frac{1}{\lambda \log \left(1+\frac{1}{\lambda \sigma^2} \right)}$.
Substituting this into~\eqref{eq:regret-bound} completes the proof. 
\end{proof-sketch}
\end{theorem}
Note that since $n$ is large in our case, assumption (a) for the above theorem is reasonable. Assumptions (b) and (c) define the upper and lower bounds on the regularization parameter $\lambda$. Similar to epoch-greedy, transferring information across the graph reduces the regret by a factor dependent on $\Tr(L^{-1})$. Note that compared to epoch-greedy, the regret bound for Thompson sampling has a worse dependence on $n$, but its $\tilde{O}(\sqrt{T})$ dependence on $T$ is optimal. If $L = I_{dn}$, we match the $\tilde{O}(dn \sqrt{T})$ regret bound for a $dn$-dimensional contextual bandit problem~\cite{abbasi2011improved}. Note that we have a dependence on $d$ and $n$ similar to the original GOB paper~\cite{cesa2013gang} and that this method performs similarly in practice in terms of regret. However, as will see, our algorithm is much faster.

%% file: Experiments.tex
\section{Experiments}
\label{sec:experiments}
\subsection{Experimental Setup} 
\textbf{Data:} We first test the scalability of various algorithms using synthetic data and then evaluate their regret performance on two real datasets. For synthetic data we generate random $d$-dimensional context vectors and ground-truth user preferences, and generate the ratings according to the linear model. We generated a random Kronecker graph with sparsity $0.005$ (which is approximately equal to the sparsity of our real datasets). It is well known that such graphs capture many properties of real-world social networks~\cite{leskovec2010kronecker}. 

For the real data, we use the Last.fm and Delicious datasets which are available as part of the HetRec 2011 workshop. Last.fm is a music streaming website where each item corresponds to a music artist and the dataset consists of the set of artists each user has listened to. The associated social network consists of $1.8$K users (nodes) and $12.7$K friendship relations (edges). Delicious is a social bookmarking website, where an item corresponds to a particular URL and the dataset consists of the set of websites bookmarked by each user. Its corresponding social network consists of $1.8$K users and $7.6$K user-user relations. Similar to~\cite{cesa2013gang}, we use the set of associated tags to construct the TF-IDF 
vector for each item and reduce the dimension of these vectors to $d = 25$. An artist (or URL) that a user has listened to (or has bookmarked) is said to be ``liked'' by the user. In each round, we select a target user uniformly at random and make the set $\Ct$ consist of $25$ randomly chosen items such that there is at least 1 item liked by the target user. An item liked by the target user is assigned a reward of $1$ whereas other items are assigned a zero reward. We use a total of $T$ = $50$ thousand recommendation rounds and average our results across $3$ runs. 

\textbf{Algorithms:} We denote our graph-based epoch-greedy and Thompson sampling algorithms as G-EG and G-TS, respectively. For epoch-greedy, although the theory suggests that we update the preference estimates only in the exploration rounds, we observed better performance by updating the preference vectors in all rounds (we use this variant in our experiments). 
We use $10\%$  of the total number of rounds for exploration, and we ``exploit" in the remaining rounds. Similar to~\cite{gentile2014online}, all hyper-parameters are set using an initial validation set of $5$ thousand rounds. The best validation performance was observed for  $\lambda = 0.01$ and $\sigma = 1$. To control the amount of exploration for Thompson sampling, we the use posterior reshaping trick~\cite{chapelle2011empirical} which reduces the variance of the posterior by a factor of $ 0.01$.

\begin{figure*}[ht]
\centering
        \subfigure[]
        {
			\includegraphics[scale = 0.39]{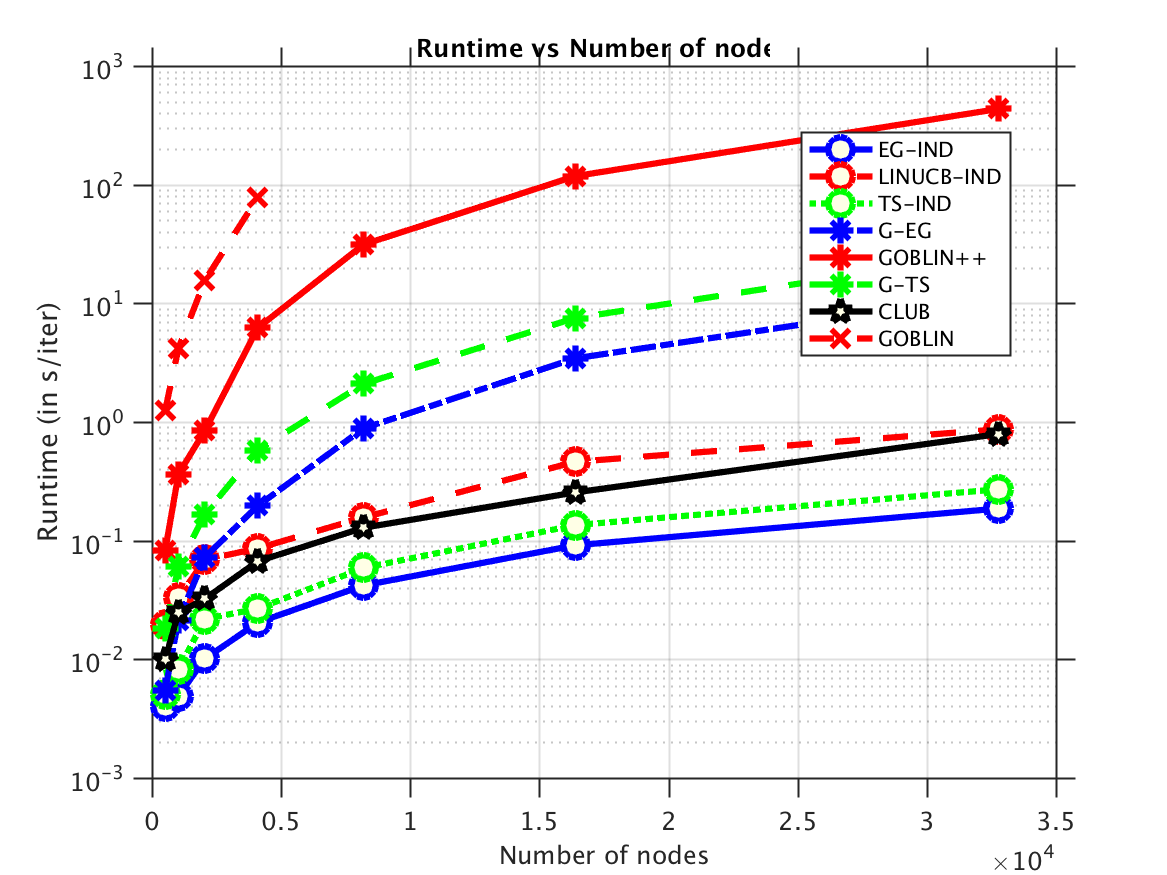}
			\label{fig:n-scalability}
        }        
        \subfigure[]
        {
			\includegraphics[scale = 0.39]{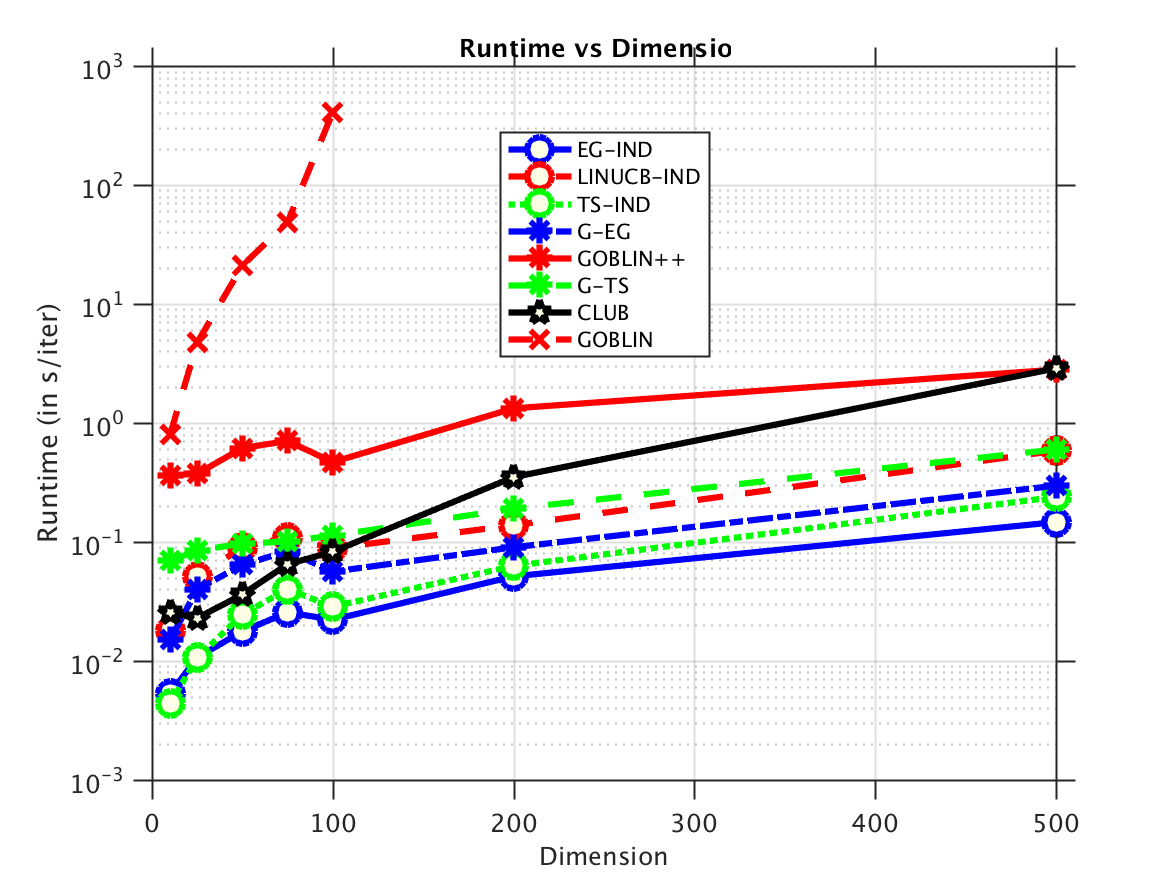}
			\label{fig:d-scalability}
        }          
        \caption{Synthetic network: Runtime (in seconds/iteration) vs (a) Number of nodes (b) Dimension}      	
\end{figure*}    
\begin{figure*}[ht]
\centering
        \subfigure[Last.fm]
        {
			\includegraphics[scale = 0.2]{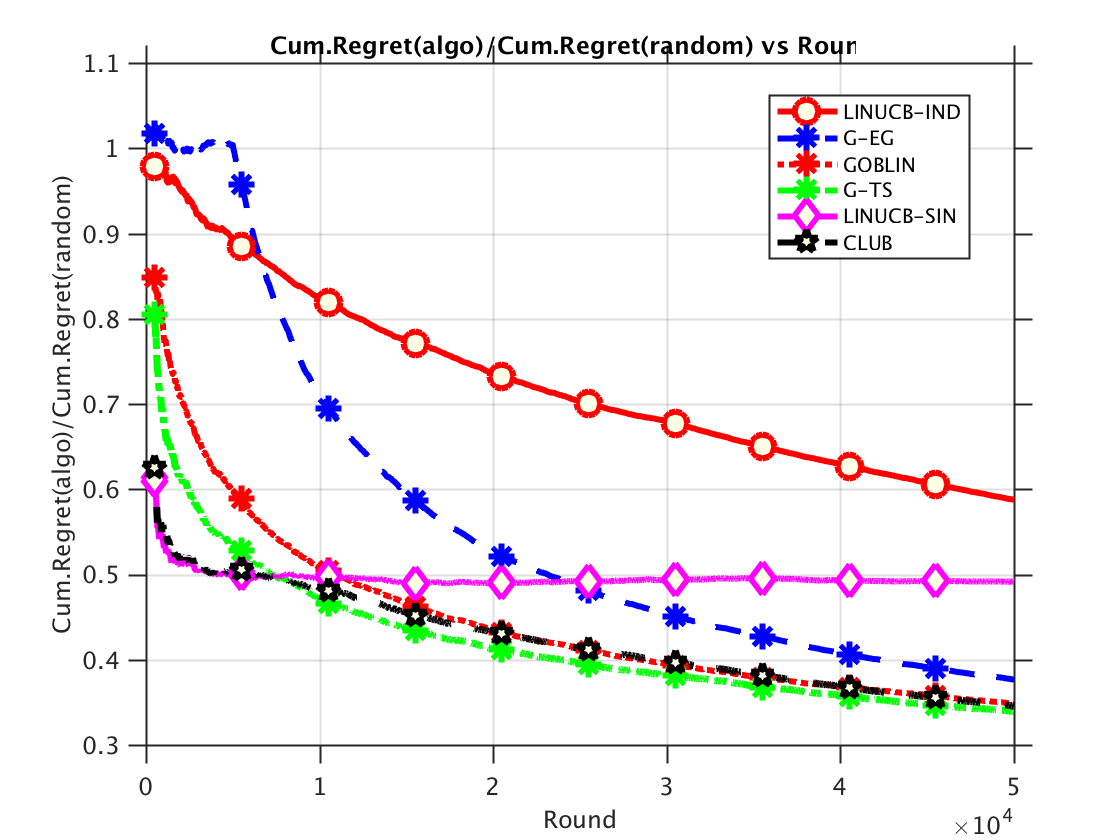}
			\label{fig:lastfm-regret}
        }
        \subfigure[Delicious]
        {
   			\includegraphics[scale = 0.2]{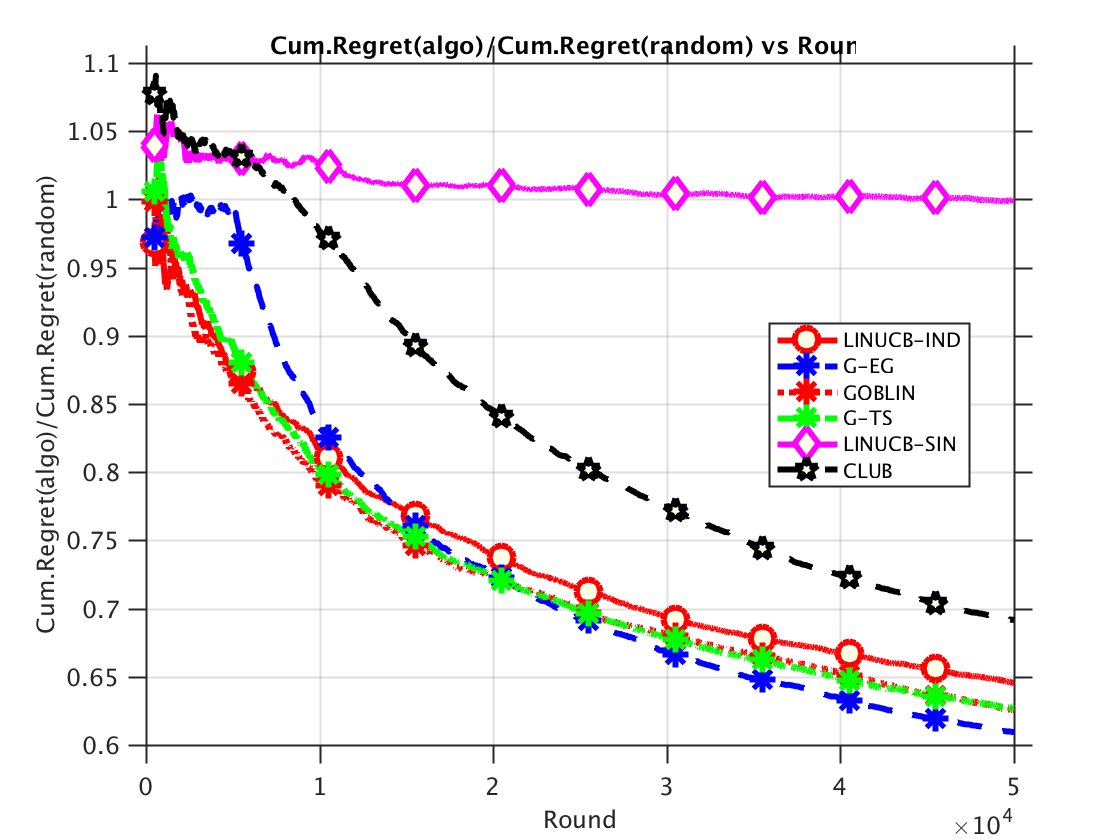}
			\label{fig:delicious-regret}
        } 
        \caption{Regret Minimization}
\end{figure*}

\textbf{Baselines:} We consider two variants of graph-based UCB-style algorithms: GOBLIN is the method proposed in the original GOB paper~\cite{cesa2013gang} while we use GOBLIN++ to refer to a variant that exploits the fast mean estimation strategy we develop in Section~\ref{subsec:scalability}. Similar to~\cite{cesa2013gang}, for both variants we discount the confidence bound term by a factor of $\alpha = 0.01$. 

We also include baselines which ignore the graph structure and make recommendations by solving independent linear contextual bandit problems for each user.  We consider 3 variants of this baseline: the LINUCB-IND proposed in~\cite{li2010contextual}, an epoch-greedy variant of this approach (EG-IND), and a Thompson sampling variant (TS-IND). We also compared to a baseline that does no personalization and simply considers a single bandit problem across all users (LINUCB-SIN). Finally, we compared against the state-of-the-art online clustering-based approach proposed in~\cite{gentile2014online}, denoted CLUB. This method starts with a fully connected graph and iteratively deletes edges from the graph based on UCB estimates. CLUB considers each connected component of this graph as a cluster and maintains one preference vector for all the users belonging to a cluster. Following the original work, we make CLUB scalable by generating a random Erdos-Renyi graph $G_{n,p}$ with $p = \frac{3 log n}{n}$.\footnote{We reimplemented CLUB. Note that one of the datasets from our experiments was also used in that work and we obtain similar performance to that reported in the original paper.} In all, we compare our proposed algorithms G-EG and G-TS with 7 reasonable baseline methods.
\subsection{Results} 
\textbf{Scalability:} We first evaluate the scalability of the various algorithms with respect to the number of network nodes $n$. Figure~\ref{fig:n-scalability} shows the runtime in seconds/iteration 
when we fix $d=25$ and vary the size of the network from $16$ thousand to $33$ thousand nodes. Compared to GOBLIN, our proposed GOBLIN++ is more efficient in terms of both time (almost 2 orders of magnitude faster) and memory. 
Indeed, the existing GOBLIN method runs out of memory even on very small networks and thus we do not plot it for larger networks.
Further, our proposed G-EG and G-TS methods scale even more gracefully in the number of nodes and are much faster than GOBLIN++ (although not as fast as the clustering-based CLUB or methods that ignore the graph). 

We next consider scalability with respect to $d$. Figure~\ref{fig:d-scalability} fixes $n=1024$ and varies $d$ from $10$ to $500$. In this figure it is again clear that our proposed GOBLIN++ scales much better than the original GOBLIN algorithm. The EG and TS variants are again even faster, and other key findings from this experiment are (i) it was not faster to ignore the graph and (ii) our proposed G-EG and G-TS methods scale better with $d$ than CLUB.
    
\textbf{Regret Minimization:} We follow~\cite{gentile2014online} in evaluating recommendation performance by plotting the ratio of cumulative regret incurred by the algorithm divided by the regret incurred by a random selection policy. Figure~\ref{fig:lastfm-regret} plots this measure for the Last.fm dataset. In this dataset we see that treating the users independently (LINUCB-IND) takes a long time to drive down the regret (we do not plot EG-IND and TS-IND as they had similar performance) while simply aggregating across users (LINUCB-SIN) performs well initially (but eventually stops making progress). We see that the approaches exploiting the graph help learn the user preferences faster than the independent approach and we note that on this dataset our proposed G-TS method performed similar to or slightly better than the state of the art CLUB algorithm.


Figure~\ref{fig:delicious-regret} shows performance on the Delicious dataset. On this dataset personalization is more important and we see that the independent method (LINUCB-IND) outperforms the non-personalized (LINUCB-SIN) approach. The need for personalization in this dataset also leads to worse performance of the clustering-based CLUB method, which is outperformed by all methods that model individual users. On this dataset the advantage of using the graph is less clear, as the graph-based methods perform similar to the independent method. Thus, these two experiments suggest that (i) the scalable graph-based methods do no worse than ignoring the graph in cases where the graph is not helpful and (ii) the scalable graph-based methods can do significantly better on datasets where the graph is helpful. Similarly, when user preferences naturally form clusters our proposed methods perform similarly to CLUB, whereas on datasets where individual preferences are important our methods are significantly better.

%% file: Conclusion.tex
\section{Discussion}
\label{sec:conclusion}
This work draws a connection between the GOB framework and GMRFs, and uses this to scale up the existing GOB model to much larger graphs. We also proposed and analyzed Thompson sampling and epoch-greedy variants. Our experiments on recommender systems datasets indicate that the Thompson sampling approach in particular is much more scalable than existing GOB methods, obtains theoretically optimal regret, and performs similar to or better than other existing scalable approaches.

In many practical scenarios we do not have an explicit graph structure available. In the supplementary material we consider a variant of the GOB model where we use L1-regularization to learn the graph on the fly. Our experiments there show that this approach works similarly to or much better than approaches which use the fixed graph structure. It would be interesting to explore the theoretical properties of this approach.

\pagebreak 


%% file: Supp-Learning.tex
\center{\textbf{Supplementary Material}}
\section{Learning the Graph} 
\label{subsec:learn-graph}
\vspace*{-2ex}

In the main paper, we assumed that the graph is known, but in practice such a user-user graph may not be available. In such a case, we explore a heuristic to learn the graph on the fly. The computational gains described in the main paper make it possible to simultaneously learn the user-preferences and infer the graph between users in an efficient manner. Our approach for learning the graph is related to methods proposed for multitask and multilabel learning in the batch setting~\cite{gonccalves2015multi,goncalves2014multi} and multitask learning in the online setting~\cite{saha2011online}. However, prior works that learn the graph in related settings only tackle problem with tens or hundreds of tasks/labels while we learn the graph and preferences across thousands of users. 

Let $V_{t} \in \mathbb{R}^{n \times n}$ be the inverse covariance matrix corresponding to the graph inferred between users at round $t$. Since zeroes in the inverse covariance matrix correspond to conditional independences between the corresponding nodes (users)~\cite{rue2005gaussian}, we use L1 regularization on $V_{t}$ for encouraging sparsity in the inferred graph. We use an additional regularization term $\Delta(V_{t} || V_{t-1}) $ to encourage the graph to change smoothly across rounds. This  encourages $V_{t}$ to be close to $V_{t-1}$ according to a distance metric $\Delta$. Following~\cite{saha2011online}, we choose $\Delta$ to be the log-determinant Bregman divergence given by $\Delta(X || Y) = \Tr(X Y^{-1}) - \log |X Y^{-1}| - dn$. If $W_{t} \in R^{d \times n} = [ \vecw_{1} \vecw_{2} \ldots \vecw_{n}]$ corresponds to the matrix of user preference estimates, the combined objective can be written as: 
\begin{flalign}
\label{eq:joint-objective-simplified}
[\vecw_{t},V_{t} ] =  \argmin_{\vecw, V} \vert \vert \vecr_{t} - \Phi_{t} \vecw \vert \vert_{2}^{2} + \Tr \big(V (\lambda W^{T} W + V_{t-1}^{-1}) \big) + \lambda_{2} \vert \vert V \vert \vert_{1} - (dn+1) \ln \vert V \vert
\end{flalign}
The first term in~\eqref{eq:joint-objective-simplified} is the data fitting term. The second term imposes the smoothness constraint across the graph and ensures that the changes in $V_{t}$ are smooth. The third term ensures that the learnt precision matrix is sparse, whereas the last term penalizes the complexity of the precision matrix. This function is independently convex in both $\vecw$ and $V$ (but not jointly convex), and we alternate between solving for $\vecw_{t}$ and $V_{t}$ in each round. With a fixed $V_t$, the $\vecw$ sub-problem is the same as the MAP estimation in the main paper and can be done efficiently. For a fixed $\vecw_{t}$, the $V$ sub-problem is given by
\begin{flalign}
V_{t} =  \argmin_{V} \Tr \big((V [\lambda \overline{W}_{t}^{T} \overline{W}_{t} + V_{t-1}^{-1}) \big) + \lambda_{2} \vert \vert V \vert \vert_{1} - (dn+1) \ln \vert V \vert 
\label{eq:omega-subproblem}
\end{flalign}
Here $\overline{W}_{t}$ refers to the mean subtracted (for each dimension) matrix of user preferences. This problem can be written as a graphical lasso problem~\cite{friedman2008sparse},  $\min_{X} \Tr(SX) + \lambda_{2} ||X||_{1} - \log |X| $, where the empirical covariance matrix $S$ is equal to $\lambda \overline{W}_{t}^{T} \overline{W}_{t} + V_{t-1}^{-1}$. We use the highly-scalable second order methods described in~\cite{hsieh2011sparse,hsieh2013big} to solve~\eqref{eq:omega-subproblem}. Thus, both sub-problems in the alternating minimization framework at each round can be solved efficiently.  

For our preliminary experiments in this direction, we use the most scalable epoch-greedy algorithm for learning the graph on the fly and denote this version as L-EG. We also consider another variant, U-EG in which we start from the Laplacian matrix $L$ corresponding to the given graph and allow it to change by re-estimating the graph according to~\eqref{eq:omega-subproblem}. Since U-EG has the flexibility to infer a better graph than the one given, such a variant is important for cases where the prior is meaningful but somewhat misspecified (the given graph accurately reflects some but not all of the user similarities). Similar to~\cite{saha2011online}, we start off with an empty graph and start learning the graph only after the preference vectors have become stable, which happens in this case after each user has received $10$ recommendations. We update the graph every $1$K rounds. For both datasets, we allow the learnt graph to contain at most $100$K edges and tune $\lambda_{2}$ to achieve a sparsity level equal to 0.05 in both cases.

\begin{figure*}[ht]
\centering

        \subfigure[Last.fm]
        {
			\includegraphics[scale = 0.2]{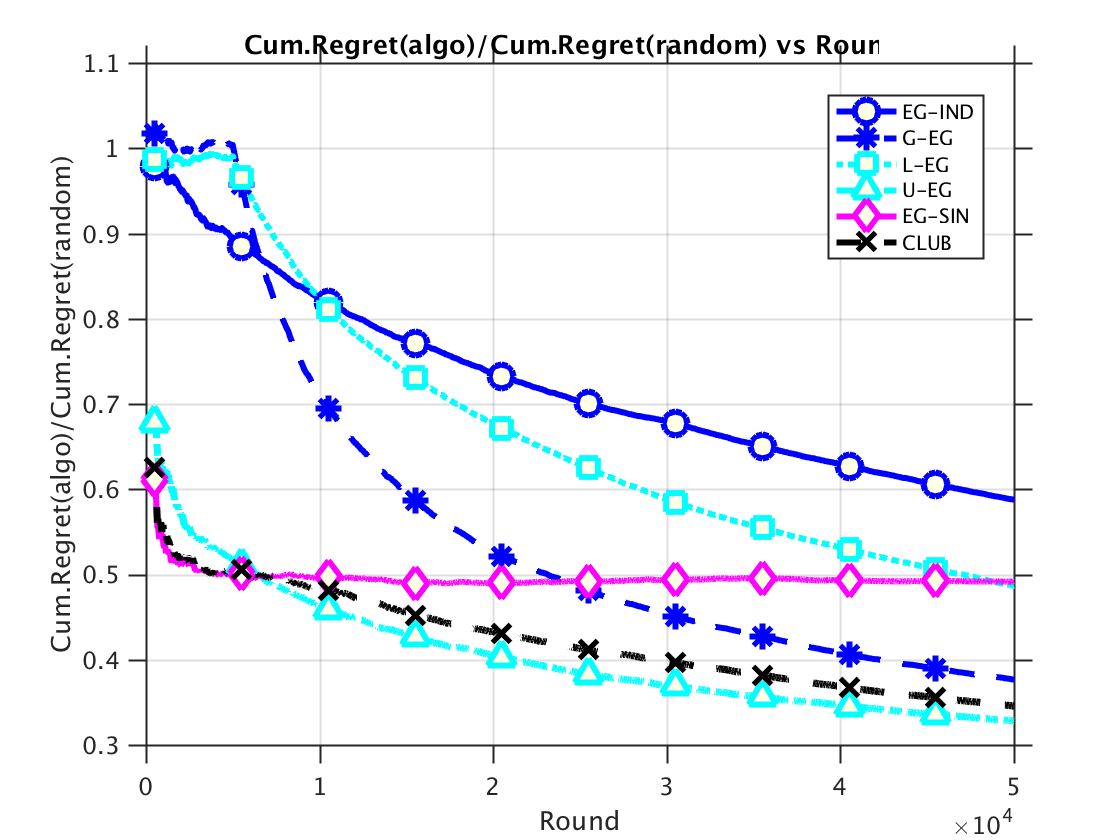}
			\label{fig:lastfm-learning-regret}
        }
        \subfigure[Delicious]
        {
   			\includegraphics[scale = 0.2]{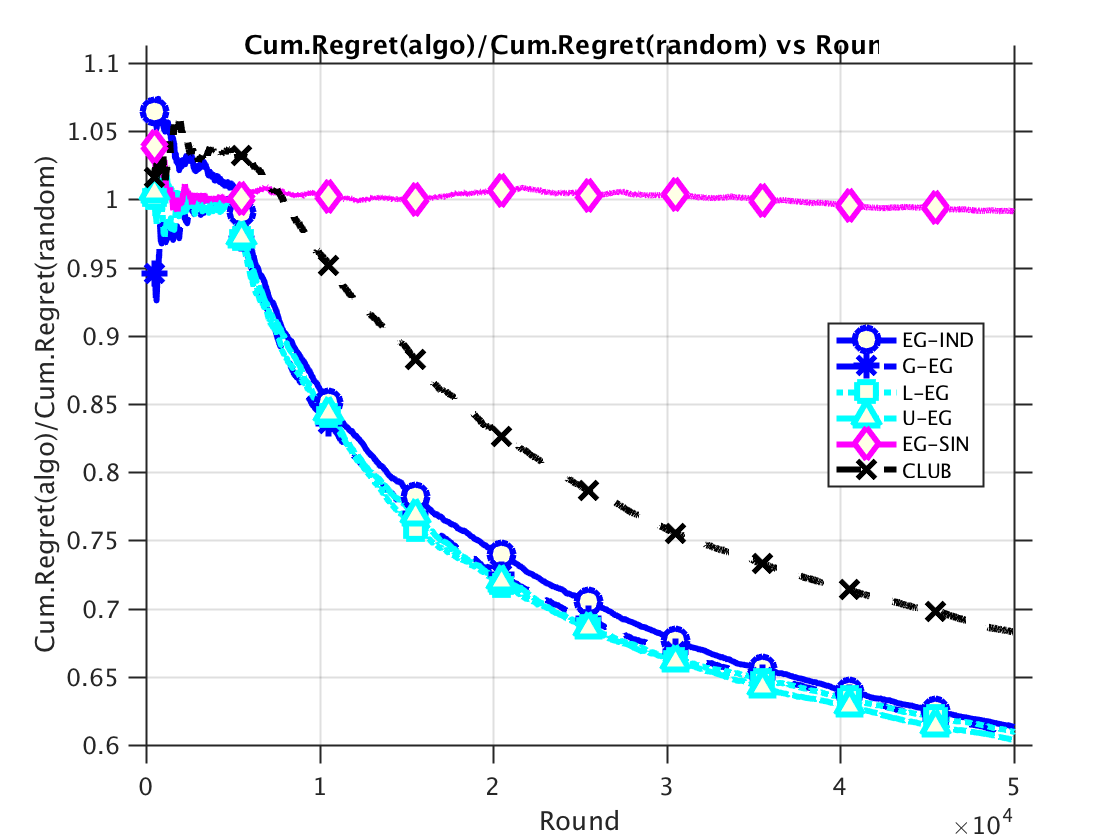}
			\label{fig:delicious-learning-regret}
        } 
        \caption{Regret Minimization while learning the graph}
\end{figure*}

To avoid clutter, we plot all the variants of the EG algorithm, L-EG and U-EG, and use EG-IND, G-EG, EG-SIN as baselines. We also plot CLUB as a baseline. For the Last.fm dataset (Figure~\ref{fig:delicious-learning-regret}(a)), U-EG performs slightly better than G-EG, which already performed well. The regret for L-EG is lower compared to LINUCB-IND indicating that learning the graph helps, but is worse as compared to both CLUB and LINUCB-SIN. On the other hand, for Delicious (Figure~\ref{fig:delicious-learning-regret}(b)), L-EG and U-EG are the best performing methods. L-EG slightly outperforms EG-IND, underscoring the importance of learning the user-user graph and transferring information between users. It also outperforms G-EG, which implies that it is able to learn a graph which reflects user similarities better than the existing social network between users. For both datasets, U-EG is among the top performing methods, which implies that allowing modifications to a good (in that it reflects user similarities reasonably well) initial graph to model the obtained data might be a good method to overcome prior misspecification. From a scalability point of view, for Delicious the running time for L-EG is $0.1083$ seconds/iteration (averaged across $T$) as compared to $0.04$ seconds/iteration for G-EG. This shows that even in the absence of an explicit user-user graph, it is possible to achieve a low regret in an efficient manner.

%% file: Supp-Proofs-Single.tex
\section{Regret bound for Epoch-Greedy}
\label{sec:EG-proof}
\renewcommand{\thetheorem}{\ref{thm:EG}}
{
\begin{theorem}
Under the additional assumption that $\vert \vert w_t \vert \vert_{2} \leq 1$ for all rounds $t$, the expected regret obtained by epoch-greedy in the GOB framework is given as:
\begin{align}
R(T) = \tilde{O}\left( n^{1/3} \left( \frac{\Tr(L^{-1})}{\lambda n} \right)^{\frac{1}{3}} T^{\frac{2}{3}} \right)
\end{align}
\begin{proof}
Let $\mathcal{H}$ be the class of hypotheses of linear functions (one for each user) coupled with Laplacian regularization. Let $\mu(\mathcal{H},q,s)$ represent the regret or cost of performing $s$ exploitation steps in epoch $q$. Let the number of exploitation steps in epoch $q$ be $s_{q}$. 

\begin{lemma}[Corollary 3.1 from~\cite{langford2008epoch}]
If $s_{q} = \lfloor \frac{1}{\mu(\mathcal{H},q,1)} \rfloor$ and $Q_{T}$ is the minimum $Q$ such that $Q+ \sum_{q = 1}^{Q} s_{q} \geq T$, then the regret obtained by Epoch Greedy is bounded by $R(T) \leq 2 Q_{T}$. 
\label{lemma:EG-basic}
\end{lemma}

We now bound the quantity $\mu(\mathcal{H},q,1)$. Let $Err(q,\mathcal{H})$ be the generalization error for $\mathcal{H}$ after obtaining $q$ unbiased samples in the exploration rounds. Clearly, 
\begin{align}
\mu(\mathcal{H},q,s) = s \cdot Err(q,\mathcal{H}).
\end{align}

Let $\ell_{LS}$ be the least squares loss. Let the number of unbiased samples per user be equal to $p$. The   empirical Rademacher complexity for our hypotheses class $\mathcal{H}$ under $\ell_{LS}$ can be given as $\hat{\mathcal{R}}_{p}^{n}(\ell_{LS} \circ \mathcal{H})$. The generalization error for $\mathcal{H}$ can be bounded as follows:
\begin{lemma}[Theorem 1 from~\cite{maurer2006rademacher}]
With probability $1 - \delta$,
\begin{align}
Err(q,\mathcal{H}) \leq \hat{\mathcal{R}}_{p}^{n}( \ell_{LS} \circ \mathcal{H}) + \sqrt{\frac{9 \ln(2/\delta)}{2pn}}
\label{eq:MTL-gen}
\end{align}
\end{lemma}

Assume that the target user is chosen uniformly at random. This implies that the expected number of samples per user is at least $p = \lfloor \frac{q}{n} \rfloor$. For simplicity, assume $q$ is exactly divisible by $n$ so that $p = \frac{q}{n}$ (this only affects the bound by a constant factor). Substituting $p$ in~\eqref{eq:MTL-gen}, we obtain
\begin{align}
Err(q,\mathcal{H}) \leq \hat{\mathcal{R}}_{p}^{n}( \ell_{LS} \circ \mathcal{H}) + \sqrt{\frac{9 \ln(2/\delta)}{2q}}.
\label{eq:MTL-gen-2}
\end{align}

The Rademacher complexity can be bounded using Lemma~\ref{lemma:EG-lemma-rademacher} (see below) as follows:
\begin{align}
\hat{\mathcal{R}}_{p}^{n}(\ell_{LS} \circ \mathcal{H}) \leq \frac{1}{\sqrt{p}}\sqrt{\frac{48 \Tr(L^{-1})}{\lambda n}} = \frac{1}{\sqrt{q}}\sqrt{\frac{48 \Tr(L^{-1})}{\lambda}}
\end{align}

Substituting this into~\eqref{eq:MTL-gen-2} we obtain
\begin{align}
Err(q,\mathcal{H}) \leq \frac{1}{\sqrt{q}} \bigg[\sqrt{\frac{48 \Tr(L^{-1})}{\lambda}} + \sqrt{\frac{9 \ln(2/\delta)}{2}} \bigg].
\label{eq:gen-err-bound}
\end{align}

We set $s_{q} = \frac{1}{Err(q,\mathcal{H})}$. Denoting $\bigg[\sqrt{\frac{48 \Tr(L^{-1})}{\lambda}} + \sqrt{\frac{9 \ln(2/\delta)}{2}} \bigg]$ as $C$, $s_{q} = \frac{\sqrt{q}}{C}$. 
\begin{align}
\intertext{Recall that from Lemma~\ref{lemma:EG-basic}, we need to determine $Q_{T}$ such that}
& Q_{T} + \sum_{q = 1}^{Q_{T}} s_{q} \geq T \implies \sum_{q = 1}^{Q_{T}} (1 + s_{q}) \geq T \nonumber
\intertext{Since $s_{q} \geq 1$, this implies that $\sum_{q = 1}^{Q_{T}} 2 s_{q} \geq T$. Substituting the value of $s_{q}$ and observing that for all $q$, $s_{q+1} \geq s_{q}$, we obtain the following:}
& 2 Q_{T} s_{Q_{T}} \geq T \implies 2 \frac{Q_{T}^{3/2}}{C} \geq T \implies Q_{T} \geq \left(\frac{CT}{2} \right)^{\frac{2}{3}} \nonumber \\
& Q_{T} = \bigg[\sqrt{\frac{12 \Tr(L^{-1})}{\lambda}} + \sqrt{\frac{9 \ln(2/\delta)}{8}} \bigg]^{\frac{2}{3}} T^{\frac{2}{3}} \label{eq:QT-bound}
\intertext{Using the above equation with Lemma~\ref{lemma:EG-basic}, we can bound the regret as}
& R(T) \leq 2 \bigg[\sqrt{\frac{12 \Tr(L^{-1})}{\lambda}} + \sqrt{\frac{9 \ln(2/\delta)}{8}} \bigg]^{\frac{2}{3}} T^{\frac{2}{3}} \label{eq:final-bound}
\intertext{To simplify this expression, we suppress the term $\sqrt{\frac{9 \ln(2/\delta)}{8}}$ in the $\tilde{O}$ notation, implying that}
& R(T) = \tilde{O}\left( 2 \bigg[\frac{12 \Tr(L^{-1})}{\lambda} \bigg]^{\frac{1}{3}} T^{\frac{2}{3}} \right) \label{eq:final-bound-2}
\end{align}
To present and interpret the result, we keep only the factors which are dependent on $n$, $\lambda$, $L$ and $T$. We then obtain 
\begin{align}
R(T) = \tilde{O}\left( n^{1/3} \left( \frac{\Tr(L^{-1})}{\lambda n} \right)^{\frac{1}{3}} T^{\frac{2}{3}} \right)
\label{eq:final-bound-3}
\end{align}
\end{proof}
\end{theorem}
\addtocounter{theorem}{-1}
}
This proves Theorem~\ref{thm:EG}. We now prove Lemma~\ref{lemma:EG-lemma-rademacher}, which was used to bound the Rademacher complexity. 

\begin{lemma}
The empirical Rademacher complexity for $\mathcal{H}$ under $\ell_{LS}$ on observing $p$ unbiased samples for each of the $n$ users can be given as:
\begin{align}
\hat{\mathcal{R}}_{p}^{n}(\ell_{LS} \circ \mathcal{H}) \leq \frac{1}{\sqrt{p}} \sqrt{\frac{48 \Tr(L^{-1})}{\lambda n}}
\end{align}
\label{lemma:EG-lemma-rademacher}
\begin{proof}
The Rademacher complexity for a class of linear predictors with graph regularization for a $0/1$ loss function $\ell_{0,1}$ can be bounded using Theorem 2 of~\cite{maurer2006rademacher}. Specifically, 
\begin{align}
\hat{\mathcal{R}}_{p}^{n}(\ell_{0,1} \circ \mathcal{H}) \leq \frac{2M}{\sqrt{p}} \sqrt{ \frac{\Tr((\lambda L)^{-1})}{n} } 
\intertext{where $M$ is the upper bound on the value of $\frac{ \vert \vert L^{\frac{1}{2}} W^{*} \vert \vert_{2}}{\sqrt{n}}$ and $W^{*}$ is the $d \times n$ matrix corresponding to the true user preferences.}
\label{eq:rademacher-classification}
\end{align}
We now upper bound $\frac{ \vert \vert L^{\frac{1}{2}} W^{*} \vert \vert_{2}}{\sqrt{n}}$. 
\begin{align}
& \vert \vert L^{\frac{1}{2}} W^{*} \vert \vert_{2} \leq \vert \vert L^{\frac{1}{2}} \vert \vert_{2} \vert \vert W^{*} \vert \vert_{2} \nonumber \\
& \vert \vert W^{*} \vert \vert_{2} \leq \vert \vert W^{*} \vert \vert_{F} = \sqrt{ \sum_{i = 1}^{n} \vert \vert w^{*}_{i} \vert \vert^{2}_{2} } \nonumber \\
& \vert \vert W^{*} \vert \vert_{2} \leq \sqrt{n} \tag{Using assumption 1: For all $i$, $\vert \vert w^{*}_{i} \vert \vert_{2} \leq 1$} \\
& \vert \vert L^{\frac{1}{2}} \vert \vert \leq {\nu_{max}(L^{\frac{1}{2}})}  = \sqrt{\nu_{max}(L)} \leq \sqrt{3} \tag{The maximum eigenvalue of any normalized Laplacian $L_{G}$ is $2$~\cite{chung1997spectral} and recall that $L = L_{G} + I_{n}$} \\
& \implies \frac{\vert \vert L^{\frac{1}{2}} W^{*} \vert \vert_{2}}{\sqrt{n}} \leq \sqrt{3} \implies M = \sqrt{3}
\end{align}

Since we perform regression using a least squares loss function instead of classification, the Rademacher complexity in our case can be bounded using Theorem 12 from~\cite{bartlett2003rademacher}. Specifically, if $\rho$ is the Lipschitz constant of the least squares problem,
\begin{align}
\hat{\mathcal{R}}_{p}^{n}( \ell_{LS} \circ \mathcal{H}) \leq 2 \rho \cdot \mathcal{R}_{p}^{n}( \ell_{0,1} \circ \mathcal{H})
\label{eq:rademacher-relation}
\end{align}
Since the estimates $w_{i,t}$ are bounded from above by $1$ (additional assumption in the theorem), $\rho = 1$. From Equations~\ref{eq:rademacher-classification},~\ref{eq:rademacher-relation} and the bound on $M$, we obtain that 
\begin{align}
\hat{\mathcal{R}}_{p}^{n}( \ell_{LS} \circ \mathcal{H}) \leq \frac{4}{\sqrt{p}} \sqrt{\frac{3 \Tr(L^{-1})}{\lambda n}}
\end{align}
which proves the lemma. 
\end{proof}
\end{lemma}

\renewcommand{\thetheorem}{\ref{thm:TS}}
{
\begin{theorem}
Under the following additional technical assumptions: (a) $\log(K) < (dn-1) \ln(2)$ (b) $\lambda < dn$ (c) $\log \left( \frac{3 + T / \lambda dn}{\delta} \right) \leq \log(KT) \log(T/\delta)$, with probability $1 - \delta$, the regret obtained by Thompson Sampling in the GOB framework is given as:
\begin{align}
R(T) = \tilde{O}\left(\frac{dn}{\sqrt{\lambda}} \sqrt{T} \sqrt{\log \left( \frac{\Tr(L^{-1})}{n} \right) + \log \left(3 + \frac{T}{\lambda dn \sigma^{2}} \right)} \right)
\end{align}
\begin{proof}
We can interpret graph-based TS as being equivalent to solving a single $dn$-dimensional contextual bandit problem, but with a modified prior covariance ($(L \otimes I_{d})^{-1}$ instead of $I_{dn}$). Our argument closely follows the proof structure in~\cite{agrawal2012thompson}, but is modified to include the prior covariance. For ease of exposition, assume that the target user at each round is implicit. We use $\kt$ to index the available items. Let the index of the optimal item at round $t$ be $\ko$, whereas the index of the item chosen by our algorithm is denoted $\ks$. 
\begin{align}
\intertext{Let $\hat{r}_{t}(\kt)$ be the estimated rating of item $\kt$ at round $t$. Then, for all $\kt$, }
& \hat{r}_{t}(\kt) \sim \mathcal{N}(\langle {\vecw}_{t}, \vecphi_{\kt} \rangle, s_{t}(\kt)) \label{eq:estimated-rating}
\intertext{Here, $s_{t}(\kt)$ is the standard deviation in the estimated rating for item $\kt$ at round $t$. Recall that $\Sigma_{t-1}$ is the covariance matrix at round $t$. $s_{t}(\kt)$ is given as:}
& s_{t}(\kt) = \sqrt{\vecphi_{\kt}^{T} \Sigma_{t-1}^{-1} \vecphi_{\kt}} \label{eq:std-def}
\intertext{We drop the argument in $s_{t}(\ks)$ to denote the standard deviation and estimated rating for the selected item $\ks$ i.e. $s_{t} = s_{t}(\ks)$ and $\hat{r}_{t} = \hat{r}_{t}(\ks)$.}
\intertext{Let $\Delta_{t}$ measure the immediate regret at round $t$ incurred by selecting item $\ks$ instead of the optimal item $\ko$. The immediate regret is given by:}
& \Delta_{t} = \langle \vecw^{*},\vecphi_{\ko} \rangle - \langle \vecw^{*}, \vecphi_{\ks} \rangle\label{eq:immediate-regret}
\intertext{Define $\cE^{\mu}(t)$ as the event such that for all $\kt$,}
& \cE^{\mu}(t): \;\; \vert \langle {\vecw}_{t}, \vecphi_{\kt} \rangle - \langle \vecw^{*}, \vecphi_{\kt} \rangle \vert \leq l_{t} s_{t}(\kt)
\intertext{Here $l_{t} = \sqrt{dn \log \left( \frac{3 + t / \lambda dn}{\delta} \right) } + \sqrt{3 \lambda}$. If the event $\cE^{\mu}(t)$ holds, it implies that the expected rating at round $t$ is close to the true rating with high probability.}
\intertext{Recall that $\vert \Ct \vert = K$ and that $\tilde{\vecw}_{t}$ is a sample drawn from the posterior distribution at round $t$. Define $\rho_{t} = \sqrt{9dn \log \left( \frac{t}{\delta} \right)}$ and $g_{t} = \min\{ \sqrt{4dn \ln(t)} , \sqrt{4 \log(tK)} \} \rho_{t} + l_{t}$. Define $\cE^{\theta}(t)$ as the event such that for all $\kt$,}
& \cE^{\theta}(t): \;\; \vert \langle \tilde{\vecw}_{t}, \vecphi_{\kt} \rangle - \langle {\vecw}_{t}, \vecphi_{\kt} \rangle \vert \leq \min\{ \sqrt{4dn \ln(t)} , \sqrt{4 \log(tK)} \} \rho_{t} s_{t}(\kt) \label{eq:e-theta}
\intertext{If the event $\cE^{\theta}(t)$ holds, it implies that the estimated rating using the sample $\tilde{\vecw}_{t}$ is close to the expected rating at round $t$.} 
\end{align}
In lemma~\ref{lemma:lemma-1}, we prove that the event $\cE^{\mu}(t)$ holds with high probability. Formally, for $\delta \in (0,1)$, 
\begin{align}
& \p(\cE^{\mu}(t)) \geq 1 - \delta
\end{align}
To show that the event $\cE^{\theta}(t)$ holds with high probability, we use the following lemma from~\cite{agrawal2012thompson}.  
\begin{lemma}[Lemma 2 of~\cite{agrawal2012thompson}]
\begin{align}
\p(\cE^{\theta}(t)) \vert \mathcal{F}_{t-1}) \geq 1 - \frac{1}{t^2}
\end{align}
\end{lemma}
Next, we use the following lemma to bound the immediate regret at round $t$. 
\begin{lemma}[Lemma 4 in~\cite{agrawal2012thompson}]
Let $\gamma = \frac{1}{4e\sqrt{\pi}}$. If the events $\cE^{\mu}(t)$ and $\cE^{\theta}(t)$ are true, then for any filtration $\Ft$, the following inequality holds:
\begin{align}
\mathbb{E}[\Delta_{t} \vert \Ft] \leq \frac{3 g_{t}}{\gamma} \mathbb{E}[s_{t} \vert \Ft] + \frac{2 g_{t}}{\gamma t^2}
\label{eq:instant-regret-bound}
\end{align}
\label{lemma:instant-regret-lemma}
\end{lemma}
\begin{align}
\intertext{Define $\cI(\cE)$ to be the indicator function for an event $\cE$. Let $regret(t) = \Delta_{t} \cdot \cI(\cE^{\mu}(t))$. We use Lemma~\ref{lemma:main-lemma} (proof is given later) which states that with probability at least $1 - \frac{\delta}{2}$,}
& \sum_{t = 1}^{T} regret(t) \leq \sum_{t = 1}^{T} \frac{3 g_{t}}{\gamma} s_{t} + \sum_{t = 1}^{T} \frac{2 g_{t}}{\gamma t^2} + \sqrt{2 \sum_{t = 1}^{T} \frac{36 g_{t}^2}{\gamma^2} \ln(2/\delta)} \label{eq:regret-bound-1} 
\intertext{From Lemma~\ref{lemma:lemma-1}, we know that event $\cE^{\mu}(t)$ holds for all $t$ with probability at least $1 - \frac{\delta}{2}$. This implies that, with probability $1 - \frac{\delta}{2}$, for all $t$ }
& regret(t) = \Delta_{t} \label{eq:inter-imm-regret}
\intertext{From Equations~\ref{eq:regret-bound-1} and~\ref{eq:inter-imm-regret}, we have that with probability $1 - \delta$,}
& R(T) = \sum_{t = 1}^{T} \Delta_{t} \leq \sum_{t = 1}^{T} \frac{3 g_{t}}{\gamma} s_{t} + \sum_{t = 1}^{T} \frac{2 g_{t}}{\gamma t^2} + \sqrt{2 \sum_{t = 1}^{T} \frac{36 g_{t}^2}{\gamma^2} \ln(2/\delta)} \nonumber 
\intertext{Note that $g_{t}$ increases with $t$ i.e. for all $t$, $g_{t} \leq g_{T}$}
& R(T) \leq \frac{3 g_{T}}{\gamma} \sum_{t = 1}^{T} s_{t} + \frac{2 g_{T}}{\gamma} \sum_{t = 1}^{T} \frac{1}{t^2} +  \frac{6g_{T}}{\gamma} \sqrt{2T \ln(2/\delta)} \label{eq:regret-bound-2} 
\intertext{Using Lemma~\ref{lemma:variance-bound} (proof given later), we have the following bound on $\sum_{t=1}^{T} s_{t}$, the variance of the selected items:}
& \sum_{t=1}^T \CB \leq \sqrt{dnT} \sqrt{C \log \left( \frac{\Tr(L^{-1})}{n} \right) + \log \left(3 + \frac{T}{\lambda dn \sigma^{2}} \right)}
\intertext{where $C = \frac{1}{\lambda \log \left(1+\frac{1}{\lambda \sigma^2} \right)}$.} 
\end{align}
Substituting this into Equation~\ref{eq:regret-bound-2}, we get 
\begin{align}
& R(T) \leq \frac{3 g_{T}}{\gamma} \sqrt{dnT} \sqrt{C \log \left( \frac{\Tr(L^{-1})}{n} \right) + \log \left(3 + \frac{T}{\lambda dn \sigma^{2}} \right)} + \frac{2 g_{T}}{\gamma} \sum_{t = 1}^{T} \frac{1}{t^2}  + \frac{6g_{T}}{\gamma} \sqrt{2T \ln(2/\delta)} \nonumber \\
\intertext{Using the fact that $\sum_{t = 1}^{T} \frac{1}{t^2} < \frac{\pi^{2}}{6}$}
& R(T) \leq \frac{3 g_{T}}{\gamma} \sqrt{dnT} \sqrt{C \log \left( \frac{\Tr(L^{-1})}{n} \right) + \log \left(3 + \frac{T}{\lambda dn \sigma^{2}} \right)}+ \frac{\pi^{2} g_{T}}{3\gamma} + \frac{6g_{T}}{\gamma} \sqrt{2T \ln(2/\delta)} \label{eq:regret-bound-3}
\end{align}
\begin{align}
\intertext{We now upper bound $g_{T}$. By our assumption on $K$, $\log(K) < (dn-1) \ln(2)$. Hence for all $t \geq 2$, $\min\{ \sqrt{4dn \ln(t)} , \sqrt{4 \log(tK)} \} = \sqrt{4 \log(tK)}$. Hence, 
}
g_{T} & = 6 \sqrt{dn \log(KT) \log(T/\delta)} + l_{T} \nonumber \\
& = 6 \sqrt{dn \log(KT) \log(T/\delta)} + \sqrt{dn \log \left( \frac{3 + T / \lambda dn}{\delta} \right)} + \sqrt{3 \lambda} \nonumber 
\intertext{By our assumption on $\lambda$, $\lambda < dn$. Hence,} 
g_{T} & \leq 8 \sqrt{dn \log(KT) \log(T/\delta)} + \sqrt{dn \log \left( \frac{3 + T / \lambda dn}{\delta} \right)} \nonumber 
\intertext{Using our assumption that $\log \left( \frac{3 + T / \lambda dn}{\delta} \right) \leq \log(KT) \log(T/\delta)$,}
g_{T} & \leq 9 \sqrt{dn \log(KT) \log(T/\delta)} \nonumber \\
\end{align}
\begin{align}
\intertext{Substituting the value of $g_{T}$ into Equation~\ref{eq:regret-bound-3}, we obtain the following:}
R(T) & \leq \frac{27 dn}{\gamma} \sqrt{T} \sqrt{C \log \left( \frac{\Tr(L^{-1})}{n} \right) + \log \left(3 + \frac{T}{\lambda dn \sigma^{2}} \right)} \nonumber \\ 
& + \frac{3 \pi^{2} \sqrt{dn \ln(T/\delta) \ln(KT)}}{\gamma} + \frac{54 \sqrt{dn \ln(T/\delta) \ln(KT)} \sqrt{2T \ln(2/\delta)}}{\gamma} \nonumber 
\end{align}
For ease of exposition, we keep the just leading terms on $d$, $n$ and $T$. This gives the following bound on $R(T)$. 
\begin{align}
R(T) = \tilde{O}\left(\frac{27 dn}{\gamma} \sqrt{T} \sqrt{C \log \left( \frac{\Tr(L^{-1})}{n} \right) + \log \left(3 + \frac{T}{\lambda dn \sigma^{2}} \right)} \right) \nonumber 
\end{align}
Rewriting the bound to keep only the terms dependent on $d$, $n$, $\lambda$, $T$ and $L$. We thus obtain the following equation. 
\begin{align}
R(T) = \tilde{O}\left(\frac{dn}{\sqrt{\lambda}} \sqrt{T} \sqrt{\log \left( \frac{\Tr(L^{-1})}{n} \right) + \log \left(3 + \frac{T}{\lambda dn \sigma^{2}} \right)} \right)\label{eq:regret-bound-5}
\end{align} 
This proves the theorem. 
\end{proof}
\end{theorem}
\addtocounter{theorem}{-1}
}
We now prove the the auxiliary lemmas used in the above proof. 

In the following lemma, we prove that $\cE^{\mu}(t)$ holds with high probability,  i.e., the expected rating at round $t$ is close to the true rating with high probability. 
\begin{lemma}
\label{lemma:lemma-1}
\begin{align}
\intertext{The following statement is true for all $\delta \in (0,1)$:}
& \Pr(\mathbb{E}^{\mu}(t)) \geq 1 - \delta
\end{align}
\begin{proof}
\begin{align}
\intertext{Recall that $r_{t} = \langle \vecw^{*}, \phi_{\ks} \rangle + \eta_{t}$ (Assumption 2) and that $\Sigma_{t} {\vecw}_{t} = \frac{b_{t}}{\sigma^{2}}$. Define $\vecs_{t-1} = \sum_{l = 1}^{t-1} \eta_{l} \vecphi_{j_{l}}$.}
& \vecs_{t-1} = \sum_{l = 1}^{t-1} \left( r_{l} - \langle \vecw^{*}, \phi_{j_{l}} \rangle \right)\vecphi_{j_{l}} = \sum_{l = 1}^{t-1} \left( r_{l} \vecphi_{j_{l}} - \vecphi_{j_{l}} \vecphi^{T}_{j_{l}} \vecw^{*}  \right) \nonumber \\
& \vecs_{t-1} = b_{t-1} - \sum_{l=1}^{t-1} \left( \vecphi_{j_{l}} \vecphi_{j_{l}}^{T} \right) \vecw^{*} = b_{t-1} - \sigma^{2}( \Sigma_{t-1} - \Sigma_{0}) \vecw^{*} = \sigma^{2}( \Sigma_{t-1} {\vecw}_{t} - \Sigma_{t-1} \vecw^{*} + \Sigma_{0} \vecw^{*}) \nonumber \\
& \hat{\vecw}_{t} - \vecw^{*} = \Sigma^{-1}_{t-1} \left( \frac{\vecs_{t-1}}{\sigma^{2}} - \Sigma_{0}  \vecw^{*} \right) \nonumber 
\end{align}
\begin{align}
\intertext{The following holds for all $\kt$:}
\vert \langle {\vecw}_{t}, \vecphi_{\kt} \rangle - \langle \vecw^{*}, \vecphi_{\kt} \rangle \vert &=
\vert \langle \vecphi_{\kt}, {\vecw}_{t} - \vecw^{*} \rangle \vert \nonumber  \\
& \leq \bigg \vert \vecphi_{\kt}^{T} \Sigma^{-1}_{t-1} \left( \frac{\vecs_{t-1}}{\sigma^{2}} - \Sigma_{0} \vecw^{*} \right) \bigg \vert \nonumber \\
& \leq \vert \vert \vecphi_{\kt} \vert \vert_{\Sigma^{-1}_{t-1}} \left( \bigg\vert \bigg\vert \frac{\vecs_{t-1}}{\sigma^{2}} - \Sigma_{0} \vecw^{*} \bigg\vert \bigg\vert_{\Sigma^{-1}_{t-1}}  \right) \tag{Since $\Sigma^{-1}_{t-1}$ is positive definite} 
\intertext{By triangle inequality,}
\vert \langle {\vecw}_{t}, \vecphi_{\kt} \rangle - \langle \vecw^{*}, \vecphi_{\kt} \rangle \vert & \leq \vert \vert \vecphi_{\kt} \vert \vert_{\Sigma_{t-1}^{-1}} \left( \bigg\vert \bigg\vert \frac{\vecs_{t-1}}{\sigma^{2}} \bigg\vert \bigg\vert_{\Sigma_{t-1}^{-1}} +  \vert \vert \Sigma_{0} \vecw^{*} \vert \vert_{\Sigma_{t-1}^{-1}} \right)\label{eq:l21} 
\end{align}
\begin{align}
\intertext{We now bound the term $\vert \vert \Sigma_{0} \vecw^{*} \vert \vert_{\Sigma_{t-1}^{-1}}$}
\vert \vert \Sigma_{0} \vecw^{*} \vert \vert_{\Sigma_{t-1}^{-1}} & \leq \vert \vert \Sigma_{0} \vecw^{*} \vert \vert_{\Sigma_{0}^{-1}} = \sqrt{ \vecw^{*T} \Sigma^{T}_{0} \Sigma^{-1}_{0} \Sigma_{0} \vecw^{*}} \tag{Since $\phi_{\ks} \phi_{\ks}^{T}$ is positive definite for all $t$} \\
& = \sqrt{ \vecw^{*T} \Sigma_{0} \vecw^{*} } \tag{Since $\Sigma_{0}$ is symmetric} \\
& \leq \sqrt{ \nu_{max} (\Sigma_{0}) } \vert \vert \vecw^{*} \vert \vert_{2} \nonumber \\ 
& \leq \sqrt{ \nu_{max} (\lambda L \otimes I_{d}) } \tag{$\vert \vert \vecw^{*} \vert \vert_{2} \leq 1$} \\
& = \sqrt{ \nu_{max} (\lambda L)} \tag{$\nu_{max}(A \otimes B) = \nu_{max}(A) \cdot \nu_{max}(B)$}\\
& \leq \sqrt{ \lambda \cdot \nu_{max}(L) } \nonumber \\
\vert \vert \Sigma_{0} \vecw^{*} \vert \vert_{\Sigma_{t-1}^{-1}} & \leq \sqrt{3 \lambda} \tag{The maximum eigenvalue of any normalized Laplacian is $2$~\cite{chung1997spectral} and recall that $L = L_{G} + I_{n}$} 
\end{align}
\begin{align}
\intertext{For bounding $\vert \vert \vecphi_{\kt} \vert \vert_{\Sigma_{t-1}^{-1}}$, note that}
& \vert \vert \vecphi_{\kt} \vert \vert_{\Sigma_{t-1}^{-1}} = \sqrt{\vecphi_{\kt}^{T} \Sigma_{t-1}^{-1}\vecphi_{\kt}} = s_{t}(\kt) \nonumber
\intertext{Using the above relations, Equation~\ref{eq:l21} can thus be rewritten as:}
& \vert \langle {\vecw}_{t}, \vecphi_{\kt} \rangle - \langle \vecw^{*}, \vecphi_{\kt} \rangle \vert \leq s_{t}(\kt) \left( \frac{1}{\sigma} \vert \vert \vecs_{t-1} \vert \vert_{\Sigma_{t-1}^{-1}} + \sqrt{3 \lambda}  \right)\label{eq:l1-1-mod}
\end{align}
To bound $\vert \vert \vecs_{t-1} \vert \vert_{\Sigma_{t-1}^{-1}}$, we use Theorem 1 from~\cite{abbasi2011improved} which we restate in our context. Note that using this theorem with the prior covariance equal to $I_{dn}$ gives Lemma 8 of~\cite{agrawal2012thompson}. 
\begin{theorem}[Theorem 1 of~\cite{abbasi2011improved}] 
For any $\delta > 0$, $t \geq 1$, with probability at least $1 - \delta$,  
\begin{align}
\vert \vert \vecs_{t-1} \vert \vert^{2}_{\Sigma_{t-1}^{-1}} & \leq 2 \sigma^{2} \log \left( \frac{\det(\Sigma_{t})^{1/2} \det(\Sigma_{0})^{-1/2}}{\delta} \right) \nonumber \\
\vert \vert \vecs_{t-1} \vert \vert^{2}_{\Sigma_{t-1}^{-1}} & \leq 2 \sigma^{2} \bigg( \log \left( \det(\Sigma_{t})^{1/2} \right) + \log \left( \det(\Sigma^{-1}_{0})^{1/2} \right) - \log(\delta) \bigg) \nonumber
\end{align}
\label{thm:yasin-martingale}
\end{theorem}
\begin{align}
\intertext{Rewriting the above equation,}
\vert \vert \vecs_{t-1} \vert \vert^{2}_{\Sigma_{t-1}^{-1}} & \leq \sigma^{2} \bigg( \log \left( \det(\Sigma_{t}) \right) + \log \left( \det(\Sigma^{-1}_{0}) \right) - 2 \log(\delta) \bigg) \nonumber \\
\intertext{We now use the trace-determinant inequality. For any $n \times n$ matrix $A$, $\det(A) \leq  \bigg(\frac{Tr(A)}{n}\bigg)^{n}$ which implies that $\log(\det(A)) \leq n \log \bigg(\frac{Tr(A)}{n}\bigg)$. Using this for both $\Sigma_{t}$ and $\Sigma^{-1}_{0}$, we obtain: }
& \vert \vert \vecs_{t-1} \vert \vert_{\Sigma_{t-1}^{-1}} \leq dn \sigma^{2} \bigg( \log \left( \bigg(\frac{\Tr(\Sigma_{t})}{dn}\bigg) \right) + \log \left( \bigg(\frac{\Tr(\Sigma^{-1}_{0})}{dn}\bigg) \right) - \frac{2}{dn}\log(\delta) \bigg)  \label{eq:l1-2} \\
\intertext{Next, we use the fact that}
& \Sigma_{t}  = \Sigma_{0} + \sum_{l = 1}^{t} \vecphi_{j_{l}} \vecphi_{j_{l}}^{T} \implies \Tr(\Sigma_{t})  \leq \Tr(\Sigma_{0}) + t \tag{Since $\vert \vert \vecphi_{j_{l}} \vert \vert_{2} \leq 1$} \nonumber 
\end{align}
\begin{align}
\intertext{Note that $\Tr(A \otimes B) = \Tr(A) \cdot \Tr(B)$. Since $\Sigma_{0} = \lambda L \otimes I_{d}$, it implies that $\Tr(\Sigma_{0}) = \lambda d \cdot \Tr(L)$. Also note that $\Tr(\Sigma^{-1}_{0}) = \Tr((\lambda L)^{-1} \otimes I_{d}) = \frac{d}{\lambda} \Tr(L^{-1})$. Using these relations in Equation~\ref{eq:l1-2},}
\vert \vert \vecs_{t-1} \vert \vert^{2}_{\Sigma_{t-1}^{-1}} & \leq dn \sigma^{2} \bigg( \log \left( \frac{\lambda d \Tr(L) + t}{dn} \right) + \log \left( \frac{\Tr(L^{-1})}{\lambda n} \right) - \frac{2}{dn}\log(\delta) \bigg) \nonumber \\
& \leq dn \sigma^{2} \bigg( \log \left( \frac{\Tr(L) \Tr(L^{-1})}{n^{2}} + \frac{t \Tr(L^{-1})}{\lambda d n^{2}}  \right) - \log(\delta^{\frac{2}{dn}}) \bigg) \tag{$\log(a) + \log(b) = \log(ab)$}  \\ 
& = dn \sigma^{2} \log \left(\frac{\Tr(L) \Tr(L^{-1})}{n^{2} \delta} + \frac{t \Tr(L^{-1})}{\lambda d n^{2} \delta}  \right) \tag{Redefining $\delta$ as $\delta^{\frac{2}{dn}}$} \\
\intertext{If $L = I_{n}$, $\Tr(L) = \Tr(L^{-1}) = n$, we recover the bound in~\cite{agrawal2012thompson} i.e. }
\vert \vert \vecs_{t-1} \vert \vert^{2}_{\Sigma_{t-1}^{-1}} & \leq dn \sigma^{2} \log \left( \frac{1 + t  /\lambda d n}{\delta}  \right) \label{eq:l1-special}  \\
\intertext{The upper bound for $\Tr(L)$ is $3n$, whereas the upper bound on $\Tr(L^{-1})$ is $n$. We thus obtain the following relation. }
\vert \vert \vecs_{t-1} \vert \vert^{2}_{\Sigma_{t-1}^{-1}} & \leq  dn \sigma^{2} \log \left( \frac{3}{\delta} + \frac{t}{\lambda d n \delta} \right) \nonumber \\
\vert \vert \vecs_{t-1} \vert \vert_{\Sigma_{t-1}^{-1}} & \leq \sigma \sqrt{dn \log \left( \frac{3 + t / \lambda dn}{\delta} \right) } \label{eq:l1-5} 
\end{align}
\begin{align}
\intertext{Combining Equations~\ref{eq:l1-1-mod} and~\ref{eq:l1-5}, we have with probability $1 - \delta$,}
& \vert \langle {\vecw}_{t}, \vecphi_{\kt} \rangle - \langle \vecw^{*}, \vecphi_{\kt} \rangle \vert \leq s_{t}(k) \left(\sqrt{dn \log \left( \frac{3 + t / \lambda dn}{\delta}\right) } + \sqrt{3 \lambda} \right) \nonumber \\
& \vert \langle {\vecw}_{t}, \vecphi_{\kt} \rangle - \langle \vecw^{*}, \vecphi_{\kt} \rangle \vert \leq s_{t}(k) l_{t} \nonumber 
\intertext{where $l_{t} = \sqrt{dn \log \left( \frac{3 + t / \lambda dn}{\delta}\right) } + \sqrt{3 \lambda}$. This completes the proof.}
\end{align}
\end{proof}
\end{lemma}
\begin{lemma}
\label{lemma:main-lemma}
With probability $1 - \delta$, 
\begin{align}
\sum_{t = 1}^{T} regret(t) \leq \sum_{t = 1}^{T} \frac{3 g_{t}}{\gamma} s_{t} + \sum_{t = 1}^{T} \frac{2 g_{t}}{\gamma t^2} + \sqrt{2 \sum_{t = 1}^{T} \frac{36 g_{t}^2}{\gamma^2} \ln{\frac{2}{\delta}}}
\end{align}
\begin{proof}
\begin{align}
\intertext{Let $Z_{l}$ and $Y_{t}$ be defined as follows:}
& Z_{l} = regret(l) - \frac{3 g_{l}}{\gamma} s_{l} - \frac{2 g_{l}}{\gamma l^2} \nonumber \\
& Y_{t} = \sum_{l = 1}^{t} Z_{l} \label{eq:Yt} \\
& \mathbb{E}[Y_{t} - Y_{t-1} \vert \Ft] = \mathbb{E}[X_{t}] = \mathbb{E}[ regret(t) \vert \Ft] - \frac{3 g_{t}}{\gamma} s_{t} - \frac{2 g_{t}}{\gamma t^2} \nonumber \\
& \mathbb{E}[regret(t) \vert \Ft] \leq \mathbb{E}[\Delta_{t}\vert \Ft] \leq \frac{3 g_{t}}{\gamma} s_{t} - \frac{2 g_{t}}{\gamma t^2} \tag{Definition of $regret(t)$ and using lemma~\ref{lemma:instant-regret-lemma}} \\
& \mathbb{E}[Y_{t} - Y_{t-1} \vert \Ft] \leq 0 \nonumber \\
\intertext{Hence, $Y_{t}$ is a super-martingale process. We now state and use the Azuma-Hoeffding inequality for $Y_{t}$}
\end{align}
\begin{definition}[Azuma-Hoeffding]
If a super-martingale $Y_{t}$ (with $t \geq 0$) and its the corresponding filtration $\Ft$, satisfies $\vert Y_{t} - Y_{t-1} \vert \leq c_{t}$ for some constant $c_{t}$, for all $t = 1, \ldots T$, then for any $a \geq 0$,
\begin{align}
Pr(Y_{T} - Y_{0} \geq a) \leq exp \left( \frac{-a^{2}}{2 \sum_{t = 1}^{T} c_{t}^{2}} \right)
\end{align}
\end{definition}

\begin{align}
\intertext{We define $Y_{0} = 0$. Note that $\vert Y_{t} - Y_{t-1} \vert = \vert Z_{l} \vert$ is bounded by $1 + \frac{3 g_{l}}{\gamma} - \frac{2 g_{l}}{\gamma l^2}$. Hence, $c_{t} = \frac{6 g_{t}}{\gamma}$. Setting $a = \sqrt{2 \ln(2/\delta) \sum_{t = 1}^{T} c_{t}^{2}}$ in the above inequality, we obtain that with probability $1 - \frac{\delta}{2}$,}
& Y_{T} \leq \sqrt{2 \sum_{t = 1}^{T} \frac{36 g_{t}^2}{\gamma^2} \ln(2/\delta)} \nonumber \\
& \sum_{t = 1}^{T} \left( regret(t) - \frac{3 g_{t}}{\gamma} s_{t} - \frac{2 g_{t}}{\gamma t^2} \right) \leq \sqrt{2 \sum_{t = 1}^{T} \frac{36 g_{t}^2}{\gamma^2} \ln(2/\delta)} \\
& \sum_{t = 1}^{T} regret(t) \leq \sum_{t = 1}^{T} \frac{3 g_{t}}{\gamma} s_{t} + \sum_{t = 1}^{T} \frac{2 g_{t}}{\gamma t^2} + \sqrt{2 \sum_{t = 1}^{T} \frac{36 g_{t}^2}{\gamma^2} \ln(2/\delta)}
\end{align}
\end{proof}
\end{lemma}

\begin{lemma}
\label{lemma:variance-bound}
\begin{align}
\sum_{t=1}^T \CB \leq \sqrt{dnT} \sqrt{C \log \left( \frac{\Tr(L^{-1})}{n} \right) + \log \left(3 + \frac{T}{\lambda dn \sigma^{2}} \right)}
\end{align}
\begin{proof}
\begin{align}
\intertext{Following the proof in~\cite{dani2008stochastic,wen2015efficient},}
\det \left[ \Sigma_t \right] & \geq \det \left[\Sigma_{t-1} + \frac{1}{\sigma^2} \vecphi_{\ks} \vecphi_{\ks}^{T} \right ] \nonumber \\
= & \det \left[ \Sigma_{t-1}^{\frac{1}{2}} \left( I + \frac{1}{\sigma^2} \Sigma_{t-1}^{-\frac{1}{2}} \vecphi_{\ks} \vecphi_{\ks}^{T} \Sigma_{t-1}^{-\frac{1}{2}} \right)\Sigma_{t-1}^{\frac{1}{2}}\right] \nonumber \\
= & \det \left[\Sigma_{t-1} \right] \det \left[  I + \frac{1}{\sigma^2} \Sigma_{t-1}^{-\frac{1}{2}} \vecphi_{\ks} \vecphi_{\ks}^{T} \Sigma_{t-1}^{-\frac{1}{2}} \right] \nonumber \\
\det \left[ \Sigma_t \right] = & \det \left[\Sigma_{t-1} \right] \left( 1+ \frac{1}{\sigma^2} \vecphi_{\ks}^{T} \Sigma_{t-1}^{-1} \vecphi_{\ks}  \right) = \det \left[\Sigma_{t-1} \right] \left( 1+ \frac{\CBsq}{\sigma^2} \right) \nonumber \\
\log \left( \det \left[ \Sigma_t \right ]\right) & \geq \log \left( \det \left[ \Sigma_{t-1} \right ]\right) + \log \left( 1+ \frac{\CBsq}{\sigma^2} \right) \nonumber \\
\log \left( \det \left[ \Sigma_T \right ]\right) & \geq \log \left( \det \left[ \Sigma_{0} \right ]\right) + \sum_{t=1}^T \log \left( 1+ \frac{\CBsq}{\sigma^2} \right) 
\label{eq:gram-determinant-recurrence}
\end{align}
\begin{align}
\intertext{If $A$ is an $n \times n$ matrix, and $B$ is an $d \times d$ matrix, then $\det[A \otimes B] = \det[A]^{d} \det[B]^{n}$. Hence,}
& \det[\Sigma_{0}] = \det[\lambda L \otimes I_{d}] = \det[ \lambda L ]^{d} \nonumber \\
& \det[\Sigma_{0}] = [\lambda^{n} \det(L)]^{d} = \lambda^{dn} [\det(L)]^{d} \nonumber \\
& \log \left( \det[\Sigma_{0}] \right) = dn \log \left( \lambda \right) + d \log \left( \det[L] \right) \label{eq:first-gram-determinant} 
\intertext{From Equations~\ref{eq:gram-determinant-recurrence} and~\ref{eq:first-gram-determinant},}
& \log \left( \det \left[ \Sigma_T \right ]\right) \geq  \left( dn \log \left( \lambda \right) + d \log \left( \det[L] \right) \right) + \sum_{t=1}^T \log \left( 1+ \frac{\CBsq}{\sigma^2} \right) 
\label{eq:gram-determinant-final} 
\intertext{We now bound the trace of $\Tr(\Sigma_{T+1})$.}
& \Tr(\Sigma_{t+1}) = \Tr(\Sigma_{t}) + \frac{\mathbf{1}}{\sigma^2} \vecphi_{\ks} \vecphi_{\ks}^{T} \implies\Tr(\Sigma_{t+1}) \leq \Tr(\Sigma_{t}) + \frac{\mathbf{1}}{\sigma^2} \tag{Since $\vert \vert \vecphi_{\ks} \vert \vert \leq 1$} \\
& \Tr(\Sigma_{T}) \leq \Tr(\Sigma_{0}) + \frac{\mathbf{T}}{\sigma^2} \nonumber \\
\intertext{Since $\Tr(A \otimes B) = \Tr(A) \cdot \Tr(B)$}
& \Tr(\Sigma_{T}) \leq  \Tr \left( \lambda (L \otimes I_{d}) \right) + \frac{T}{\sigma^{2}} \implies \Tr(\Sigma_{T}) \leq \lambda d \Tr(L) + \frac{T}{\sigma^{2}} \label{eq:gram-trace-final} 
\intertext{Using the determinant-trace inequality, we have the following relation:}
& \left( \frac{1}{dn} \Tr(\Sigma_{T}) \right)^{dn} \geq (\det[\Sigma_{T}]) \nonumber \\ 
& dn \log \left( \frac{1}{dn} \Tr(\Sigma_{T}) \right) \geq \log \left( \det[\Sigma_{T}] \right)  \label{eq:determinant-trace-inequality} 
\intertext{Using Equations~\ref{eq:gram-determinant-final},~\ref{eq:gram-trace-final} and~\ref{eq:determinant-trace-inequality}, we obtain the following relation.}
& dn \log \left( \frac{\lambda d \Tr(L) + \frac{T}{\sigma^{2}}}{dn} \right) \geq \left( dn \log \left( \lambda \right) + d \log \left( \det[L] \right) \right) + \sum_{t=1}^T \log \left( 1+ \frac{\CBsq}{\sigma^2} \right) 
 \nonumber 
\end{align}
\begin{align}
\sum_{t=1}^T \log \left( 1+ \frac{\CBsq}{\sigma^2} \right) & \leq dn \log \left( \frac{\lambda d \Tr(L) + \frac{T}{\sigma^{2}}}{dn} \right) - dn \log \left( \lambda \right) - d \log \left( \det[L] \right) \nonumber \\
& \leq dn \log \left( \frac{\lambda d \Tr(L) + \frac{T}{\sigma^{2}}}{dn} \right) - dn \log \left( \lambda \right) + d \log \left( \det[L^{-1}] \right) \tag{$det[L^{-1}] = 1/det[L]$} \\
& \leq dn \log \left( \frac{\lambda d \Tr(L) + \frac{T}{\sigma^{2}}}{dn} \right) - dn \log \left( \lambda \right) + dn \log \left( \frac{1}{n} \Tr(L^{-1}) \right) \tag{Using the determinant-trace inequality for $\log(\det[L^{-1}])$} \\
& \leq dn \log \left( \frac{\lambda d \Tr(L) \Tr(L^{-1}) + \frac{\Tr(L^{-1}) T}{\sigma^{2}}}{\lambda dn^{2}} \right) \tag{$\log(a) + \log(b) = \log(ab)$} \\
& \leq dn \log \left( \frac{\Tr(L) \Tr(L^{-1})}{ n^{2}} + \frac{\Tr(L^{-1}) T}{\lambda dn^{2}\sigma^{2}} \right) \nonumber \\
\intertext{The maximum eigenvalue of any Laplacian is $2$. Hence $\Tr(L)$ is upper-bounded by $3n$.}
\sum_{t=1}^T \log \left( 1+ \frac{\CBsq}{\sigma^2} \right) & \leq dn \log \left( \frac{3 \Tr(L^{-1})}{n} + \frac{\Tr(L^{-1}) T}{\lambda dn^{2}\sigma^{2}} \right) \label{eq:inter-1} \\
\end{align}
\begin{align}
\CBsq & =  \vecphi_{\kt}^{T} \Sigma_{t}^{-1} \vecphi_{\kt} \leq \vecphi_{\kt}^{T} \Sigma_{0}^{-1} \vecphi_{\kt} \tag{Since we are making positive definite updates at each round $t$} \\
& \leq \|\vecphi_{\kt}\|^2 \nu_{max}( \Sigma^{-1}_{0} ) \nonumber \\
& = \|\vecphi_{\kt}\|^2 \frac{1}{\nu_{min}( \lambda L \otimes I_{d} )} \nonumber \\
& = \|\vecphi_{\kt}\|^2 \frac{1}{\nu_{min}(\lambda L)} \tag{$\nu_{min}(A \otimes B) = \nu_{min}(A) \nu_{min}(B)$}  \\
& \leq \frac{1}{\lambda} \cdot \frac{1}{\nu_{min}(L)} \tag{$\vert \vert \phi_{\kt} \vert \vert_{2} \leq 1$} \\
\CBsq & \leq \frac{1}{\lambda} \tag{Minimum eigenvalue of a normalized Laplacian $L_{G}$ is $0$. $L = L_{G} + I_{n}$}
\end{align}
Moreover, for all $y \in [0, 1/\lambda]$, we have $\log \left(1+\frac{y}{\sigma^2} \right) \geq \lambda \log \left(1+\frac{1}{\lambda \sigma^2} \right) y$ based on the concavity of $\log (\cdot)$. To see this, consider the following function:
\begin{align}
h(y) = \frac{\log \left(1+\frac{y}{\sigma^2} \right)}{\lambda \log \left(1+\frac{1}{\lambda \sigma^2} \right)} -  y
\end{align}
Clearly, $h(y)$ is concave. Also note that, $h(0) = h(1/\lambda) = 0$. Hence for all $y \in [0, 1/\lambda]$, the function $h(y) \geq 0$. This implies that $\log \left(1+\frac{y}{\sigma^2} \right) \geq \lambda \log \left(1+\frac{1}{\lambda \sigma^2} \right) y$. We use this result by setting $y = \CBsq$. 
\begin{align}
& \log \left(1+\frac{\CBsq}{\sigma^2} \right) \geq \lambda \log \left(1+\frac{1}{\lambda \sigma^2} \right) \CBsq \nonumber \\
& \CBsq  \leq \frac{1}{\lambda \log \left(1+\frac{1}{\lambda \sigma^2} \right)} \log \left(1+\frac{\CBsq}{\sigma^2} \right) \label{eq:log-concavity-1} 
\intertext{Hence,}
& \sum_{t=1}^T \CBsq \leq \frac{1}{\lambda \log \left(1+\frac{1}{\lambda \sigma^2} \right)} \sum_{t=1}^T \log \left(1+\frac{\CBsq}{\sigma^2} \right) \label{eq:log-concavity-2} \\
\intertext{By Cauchy Schwartz,}
& \sum_{t=1}^T \CB \leq \sqrt{T} \sqrt{\sum_{t=1}^T \CBsq} \label{eq:cauchy-schwartz-1} \\
\intertext{From Equations~\ref{eq:log-concavity-2} and~\ref{eq:cauchy-schwartz-1},}
& \sum_{t=1}^T \CB \leq \sqrt{T} \sqrt{\frac{1}{\lambda \log \left(1+\frac{1}{\lambda \sigma^2} \right)} \sum_{t=1}^T \log \left(1+\frac{\CBsq}{\sigma^2} \right)} \nonumber \\
& \sum_{t=1}^T \CB \leq \sqrt{T} \sqrt{C \sum_{t=1}^T \log \left(1+\frac{\CBsq}{\sigma^2} \right)} \label{eq:inter-2} \\
\intertext{where $C= \frac{1}{\lambda \log \left(1+\frac{1}{\lambda \sigma^2} \right)}$. Using Equations~\ref{eq:inter-1} and~\ref{eq:inter-2},}
& \sum_{t=1}^T \CB \leq \sqrt{dnT} \sqrt{C \log \left( \frac{3 \Tr(L^{-1})}{n} + \frac{\Tr(L^{-1}) T}{\lambda dn^{2}\sigma^{2}} \right)} \nonumber \\
& \sum_{t=1}^T \CB \leq \sqrt{dnT} \sqrt{C \log \left( \frac{\Tr(L^{-1})}{n} \right) + \log \left(3 + \frac{T}{\lambda dn \sigma^{2}} \right)}
\end{align}
\end{proof}
\end{lemma}